%% file: latex/acl_latex.tex
\documentclass[11pt]{article}

\usepackage[final]{acl}
\usepackage{graphicx}
\usepackage{times}
\usepackage{latexsym}

\usepackage[T1]{fontenc}

\usepackage[utf8]{inputenc}

\usepackage{microtype}

\usepackage{inconsolata}

\usepackage{graphicx}
\usepackage{tcolorbox}
\usepackage{cuted} 
\usepackage{booktabs}   
\usepackage{siunitx}    
\usepackage{multirow}   
\usepackage{graphicx} 

\usepackage{booktabs}   
\usepackage{siunitx}    
\usepackage{multirow}   
\usepackage{graphicx}   

\usepackage{booktabs}   
\usepackage{multirow}   
\usepackage{siunitx}    

\usepackage{pifont}       
\usepackage{bbding}       
\definecolor{mygreen}{HTML}{1E8449} 
\definecolor{myred}{HTML}{C0392B}  
\usepackage{xcolor}     
\usepackage{colortbl}  
\usepackage{algorithm}
\usepackage{algorithmic}
\usepackage{amsfonts} 
\usepackage{amsmath}
\usepackage{booktabs}
\usepackage{siunitx}
\usepackage{multirow}

\usepackage{tcolorbox}
\title{Measure Twice, Click Once: Co-evolving Proposer and Visual Critic via Reinforcement Learning for GUI Grounding}
\author{
  \textbf{Wenkai Wang\textsuperscript{1}},
  \textbf{Xiyun Li\textsuperscript{2}},
  \textbf{Hongcan Guo\textsuperscript{3}},
  \textbf{Wenhao Yu\textsuperscript{2}},
  \textbf{Tianqing Fang\textsuperscript{2}}
\\
  \textbf{Haitao Mi\textsuperscript{2}},
  \textbf{Dong Yu\textsuperscript{2}},
  \textbf{Shengyu Zhang\textsuperscript{1} \thanks{Corresponding author.}}
\\
  \textsuperscript{1}Zhejiang University,
  \textsuperscript{2}Tencent AI Lab,
  \textsuperscript{3}The University of Hong Kong
\\
  \texttt{\{} 
  \href{mailto:wenkaiwang@zju.edu.cn}{\texttt{wenkaiwang}}, 
  \href{mailto:sy_zhang@zju.edu.cn}{\texttt{sy\_zhang}} 
  \texttt{\}@zju.edu.cn} 
}

\begin{document}
\maketitle
\begin{abstract}

\input{tex/abstract}

\end{abstract}

\begin{figure}[h]  
    \centering
    \includegraphics[width=0.95 \linewidth]{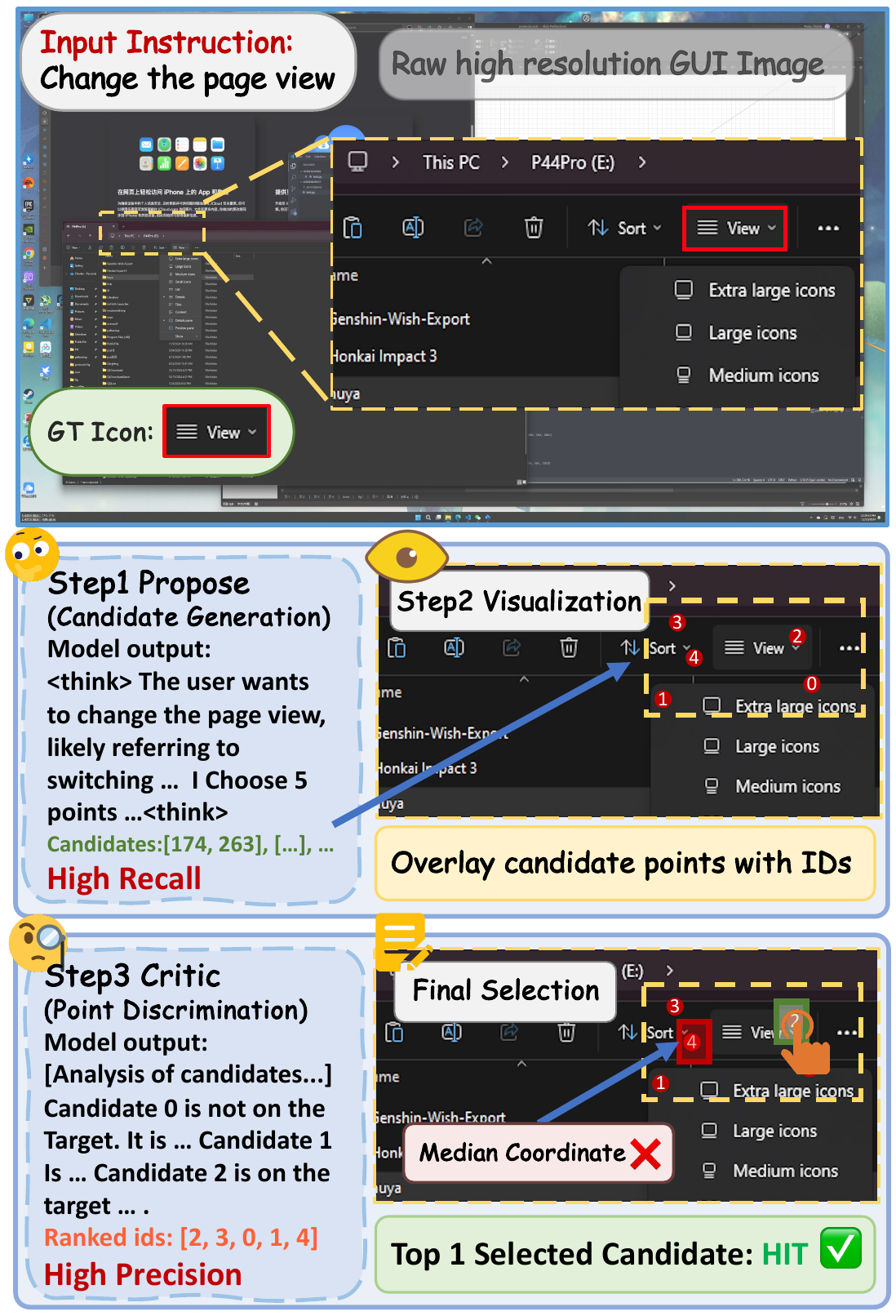} 
    \caption{\textbf{Overview of the Propose-then-Critic framework.} The process involves three steps: (1) Propose: generating diverse candidates for coverage; (2) Visualize: rendering candidates with ID markers; and (3) Critic: ranking candidates to select the precise target.}
    \label{fig:example_label}
\end{figure}

\section{Introduction}
\input{tex/intro}

\section{Related Work}
\input{tex/related_works}
\section{Motivation}
\input{tex/motivation}
\section{Methodology}
\input{tex/method}
\section{Experiment}
\input{tex/experiment}

\section{Conclusion}
\input{tex/conclusion}
\section*{Limitations}
\input{tex/limitations}
\section*{Acknowledgements}
\input{tex/acknowledgements}

\bibliography{custom}

\appendix

\input{tex/appendix}

\end{document}

%% file: tex/abstract.tex
Graphical User Interface (GUI) grounding requires mapping natural language instructions to precise pixel coordinates. However, due to visually homogeneous elements and dense layouts, models typically grasp semantic intent yet struggle with achieving precise localization. While scaling sampling attempts (Pass@k) reveals potential gains, static self-consistency strategies derived from geometric clustering often yield limited improvements, as the model's predictions tend to be spatially dispersed. 
In this paper, we propose replacing static consistency strategies with a learnable selection mechanism that selects the optimal target by critiquing its own proposals rendered on the screenshot. 
Given the significant disparity between the model's grounding and critiquing capabilities, we propose a co-evolving Propose-then-Critic framework. 
To jointly optimize these, we introduce a maturity-aware adaptive co-evolutionary reinforcement learning paradigm. This approach dynamically balances the training objectives of proposer and critic, where the diversity of the proposer's outputs enhances critic robustness, while the critic's maturing discrimination capability conversely unlocks the proposer's potential for extensive spatial exploration, fostering the mutual reinforcement and co-evolution of both capabilities, thereby ensuring generalizability to adapt to diverse and complex interface layouts. Extensive experiments over 6 benchmarks show that our method significantly enhances both grounding accuracy and critic reliability. 

%% file: tex/intro.tex
\label{sec:abstract}

The rapid evolution of autonomous Graphical User Interface (GUI) agents is reshaping the landscape of digital automation~\cite{guisurvey,guisurvey2}. At the heart of this interaction lies GUI Grounding, a fundamental capability that translates abstract natural language instructions into precise executable coordinates. Recent advancements in Multimodal Large Language Models (MLLMs) \cite{Qwen3-VL} have shown significant promise, leveraging robust vision-language alignment to interpret screens and user intents.


However, applying MLLMs to GUI Grounding remains challenging, as it demands not only semantic understanding of the user intent and screen layout but also the regression of pixel-perfect coordinates ~\cite{infixg1}. This is particularly difficult in modern GUI environments characterized by high resolutions and densely packed, small UI elements, where models often struggle to accurately localize targets in a single attempt. 

Empirically, sampling multiple candidates yields a significantly higher Pass@K than single-shot generation, suggesting that the correct target often lies within the model's potential output distribution. Nevertheless, as shown in Figure \ref{fig:Qwen3vlcopm} naive self-consistency strategies derived from geometric clustering (e.g. Geometric Median or Medoid) often yield limited improvements ~\cite{du2025rcpo}. This is largely due to stochastic dispersion of predicted coordinates, which prevents simple geometric aggregation from forming a reliable consensus.


Current methods predominately adopt a direct regression paradigm, training models to deterministically map semantic instructions to specific coordinates ~\cite{guiactor, segui, aguvis}. However, such deterministic supervision may constrain generative diversity, potentially hindering robust generalization across varied layouts ~\cite{chu2025sft}. Some training-free methods attempt to leverage output consistency ~\cite{guifree1, guifree2}, yet the high 
spatial variance renders these improvements marginal. Other works explore iterative refinement through self-reflection to improve precision ~\cite{guicursor, reflectgui, reflect2gui}. However, we observe that the critic capability of existing models typically lags behind their generative performance, limiting their effectiveness in refining coordinate proposals. \textbf{This leads to a key research question:}
\begin{tcolorbox}[colback=lightgray!10, colframe=lightgray!50, sharp corners=south, boxrule=0.5mm, left=3mm, right=3mm, top=3mm, bottom=3mm]
\textit{How can we design a learnable paradigm to effectively exploit the latent potential within the model's multiple sampling attempts?}
\end{tcolorbox}


To address this challenge, we introduce a novel Propose-then-Critic framework, which shifts GUI grounding from a single-pass regression task to a learnable Visual Perception Ranking paradigm. Instead of relying on fragile geometric clustering to aggregate disparate coordinates, we leverage the model's inner discriminative capabilities to verify its own proposals through visual feedback rendering potential targets on the screenshot to aligning semantic intent with pixel-level execution. To bridge the inherent disparity between the model's grounding and critiquing capabilities, we propose a Co-evolutionary Reinforcement Learning strategy. Unlike standard RL approaches that often stifle exploration in favor of exploiting high-reward modes, our method treats the Proposer and Critic as symbiotic agents. Crucially, we introduce a maturity-aware mechanism to harmonize the intrinsic tension between prediction accuracy and candidate diversity. This mechanism orchestrates an adaptive curriculum: it initially guides agents to master fundamental localization, and progressively evolves to encourage expansive spatial exploration as they mature. By dynamically rebalancing the learning focus, the diversity of the proposer's outputs serves to enhance the robustness of the critic, while the critic's maturing discrimination capability empowers the proposer to expand its spatial search scope, fostering a mutual reinforcement loop. This ensures the model exploits the full potential of test-time visual search without collapsing into a repetitive solution. Distinct from self-consistency methods, our framework generates multiple candidates in a single inference pass, substantially reducing token overhead compared to repeated sampling. Our contributions are as follows:

\begin{itemize}
\item We propose a Propose-then-Critic framework that transforms GUI grounding into a Visual Perception Ranking paradigm, utilizing visual feedback to bridge the instruction-visual alignment gap.

\item  We introduce a Co-evolutionary Reinforcement Learning strategy driven by a dynamic maturity mechanism, which adaptively balances prediction accuracy and diversity.

\item We achieve a relative improvement of grounding capability up to 17.2\%. Our method substantially improves both the generation accuracy and the critic capabilities of MLLMs. 


\end{itemize}



%% file: tex/related_works.tex
\subsection{GUI Grounding}
GUI Grounding aligns user instructions with specific screen elements by predicting precise execution coordinates ~\cite{cheng2024seeclick}. Mainstream MLLMs use Supervised Fine-Tuning (SFT) for direct regression, enhancing capabilities via data scaling ~\cite{cheng2024seeclick, aguvis, uground, osworldg, hong2024cogagent}. Some research explores training-free paradigms leverage test-time scaling (Sample-$K$) or self-reflection for refinement ~\cite{dimo, reflectgui, reflect2gui}. However, since SFT based methods may constrains generation diversity and static aggregation struggles with spatial variance, we propose a Reinforcement Learning framework to unlock the model's latent sampling potential.
\subsection{MLLM Reinforcement Learning}
Reinforcement Learning (RL) has been broadly leveraged to enhance visual reasoning ~\cite{guo2025deepseek, huang2025visionr1, zhang2025r1vl, li2026verified, yu2025guided,fang2025webevolver}. In the domain of GUI grounding, recent works have adopted this paradigm to either optimize reward modeling for direct regression or transform the inference paradigm ~\cite{infixg1,yang2025gta1, guig2, kang2025guirlvg,lu2025uir1}. Specifically, RCPO leverage region consistency ~\cite{du2025rcpo} as a reward signal, while other notable approaches like GUI Spotlight ~\cite{guispotlight} and GUI-Cursor ~\cite{guicursor} employ RL to enable iterative refinement. However, such dependence on self-verification may lead to erroneous corrections due to potential semantic blindness. In comparison, our maturity-aware co-evolutionary paradigm synergistically evolves both generation and discrimination capabilities, fostering robust grounding performance.

%% file: tex/motivation.tex
\begin{figure}[htbp]
  \includegraphics[width=0.47\textwidth]{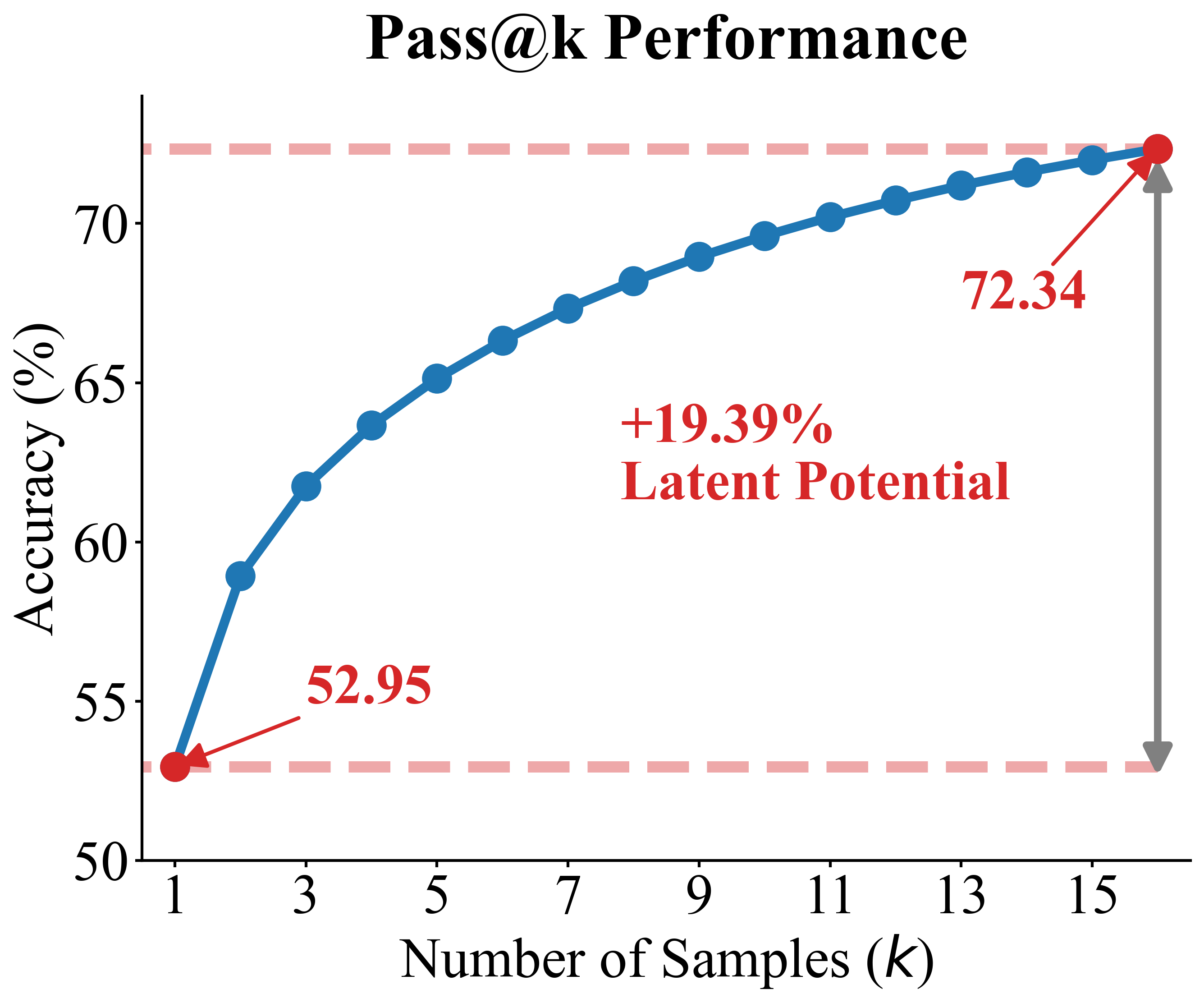}
  \caption{Analysis of Latent Potential via Pass@k. }
  \label{fig:motivation}
\end{figure}
Current GUI agents typically employ a "direct regression" paradigm, mapping instructions directly to pixel coordinates. However, this method suffers from a significant visual-semantic gap: translating abstract intent into precise pixels introduces unavoidable spatial variance. This creates a bottleneck for single-shot performance, prompting us to look beyond deterministic regression and explore  multiple sampling strategies to mitigate the high spatial variance inherent in single-shot estimation.


\textbf{The Latent Potential and Continuous Consistency Challenge. } We observe that while the single-shot accuracy (Pass@1) of GUI agents is limited, their Pass@k performance is substantially higher, indicating that the correct semantic understanding is often present but distributed across a probabilistic output space. To exploit this, one might apply Test-Time Compute techniques like Self-Consistency. However, unlike discrete text tokens, the high spatial variance and stochastic dispersion inherent in continuous coordinates prevent geometric clustering from forming a reliable consensus, leading to imprecise localization.


\textbf{The Pitfalls of Naive Visual Criticism} To bridge this gap, a natural progression is to transform the regression problem into a visual discrimination task—rendering candidate points on the screen for a VLM to judge. However, as shown in Figure 3, we observe that while off-the-shelf VLMs can easily filter out random noise, they struggle to distinguish the target from model-generated candidates, which typically cluster near the target or land on valid UI elements, acting as plausible distractors that confuse the critic.
\begin{figure}[htbp]
  \centering
  \begin{minipage}[c]{0.234\textwidth}
    \includegraphics[width=\textwidth]{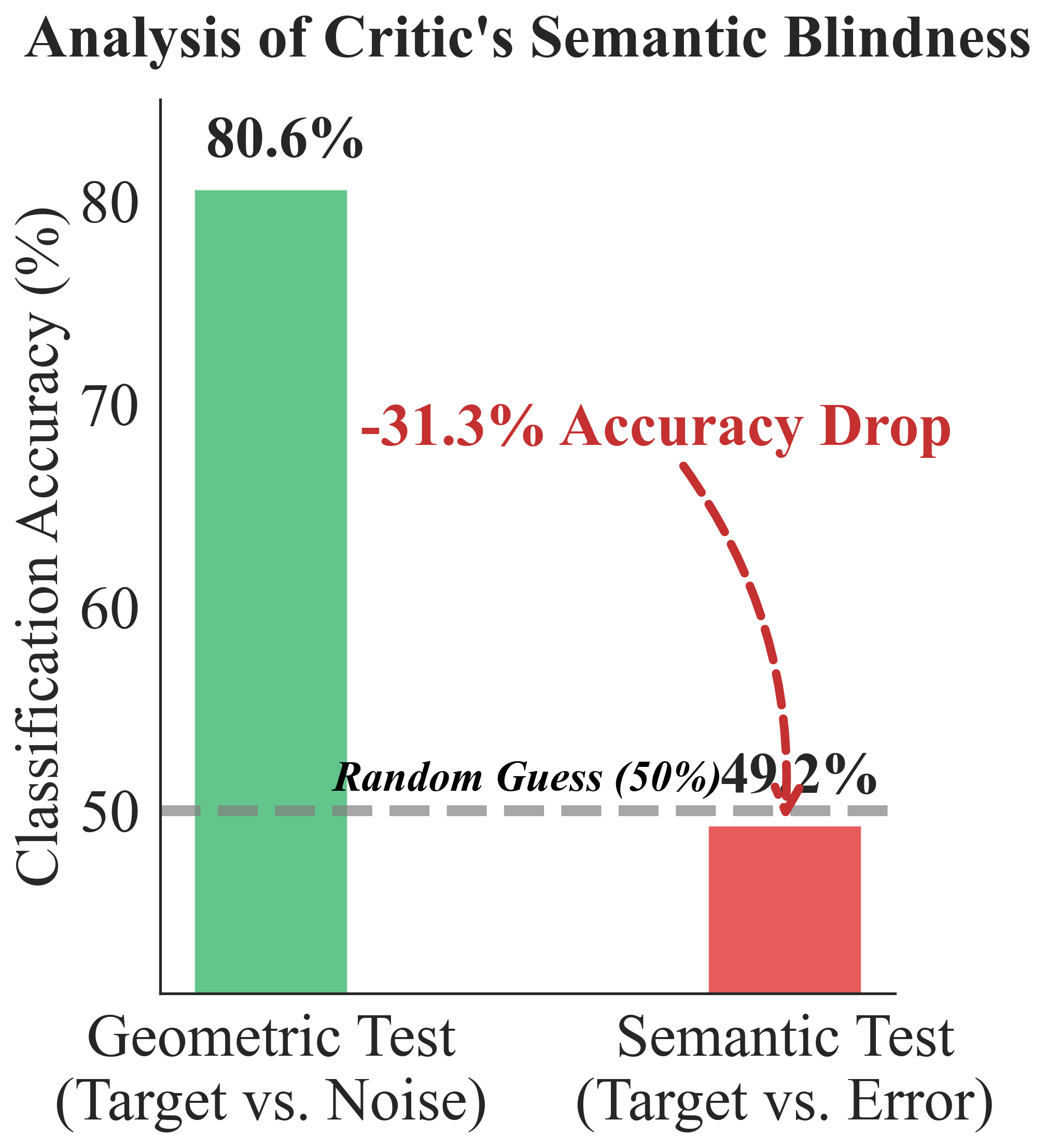}
  \end{minipage}
  \begin{minipage}[c]{0.234\textwidth}
    \includegraphics[width=\textwidth]{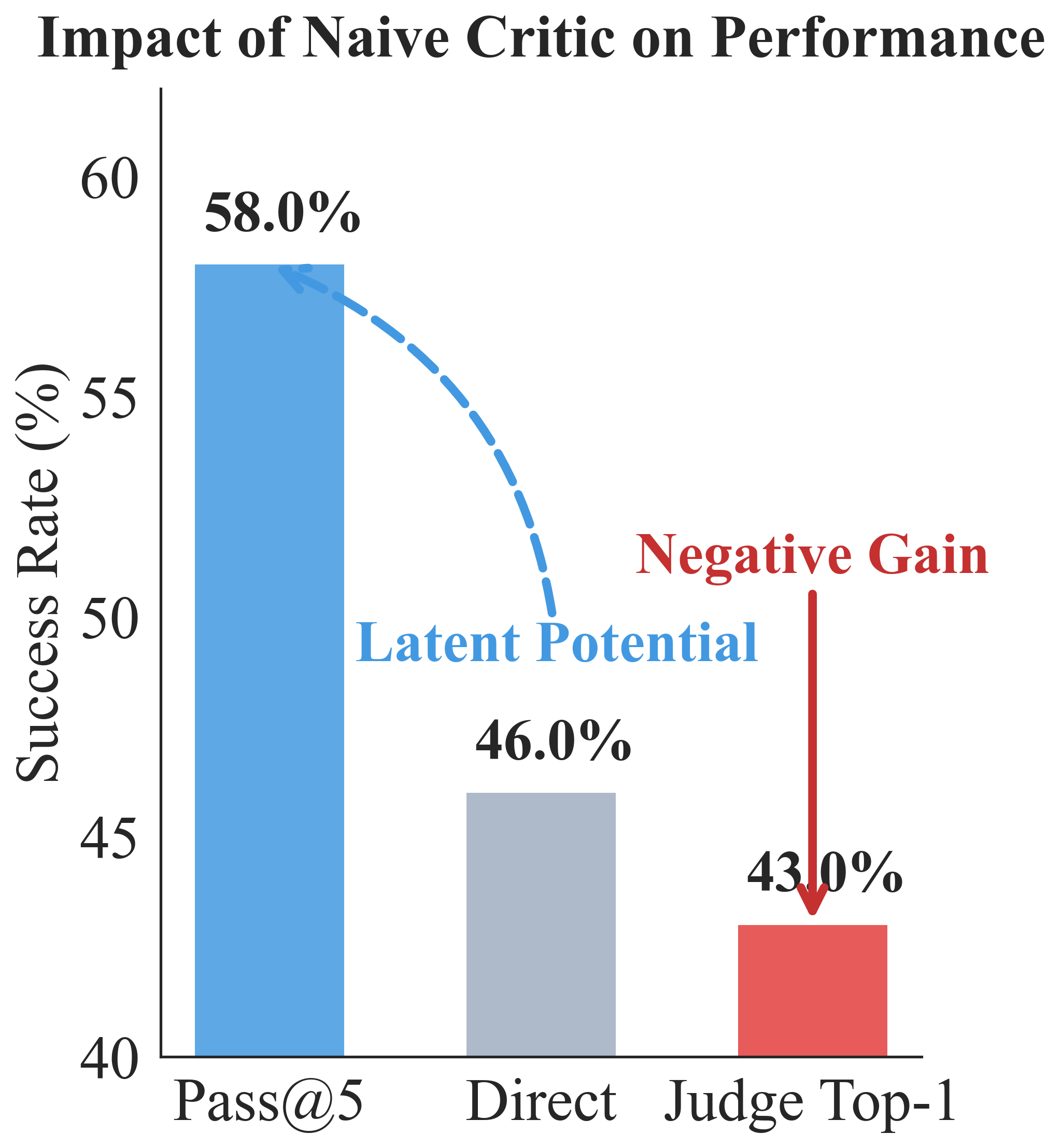}
  \end{minipage}

  \caption{Analysis of Critic's Semantic Blindness and its Negative Impact.}
  \label{fig_Variation}
\end{figure}

%% file: tex/method.tex
\begin{figure*}[htbp]
  \centering
  \includegraphics[width=\textwidth]{figure/new_method_9.pdf}
  \caption{\textbf{The Co-evolutionary Reinforcement Learning framework.} \textbf{Left:} The unified 'Propose-then-Judge' pipeline. \textbf{Middle:} Decoupled reward formulation optimizing the Proposer and the Critic. \textbf{Right:} The Maturity-Aware mechanism which dynamically schedules optimization weights to synergistically evolve generation and discrimination capabilities. } 
  \label{fig:wide}
\end{figure*}
To unlock the latent potential in multiple sampling and bridge the gap between model's generative and discriminatory capabilities, we propose the \textbf{Co}-evolutionary Framework of \textbf{P}roposer and Visual \textbf{C}ritic based on reinforcement learning (COPC). In this section, we will elaborate on the methodology of COPC, which is built upon Grouped Reward Policy Optimization (GRPO). The framework first generates the Proposer and Critic trajectories, then assigns decoupled rewards to each role. By dynamically modulating the reward weights based on training maturity, our approach facilitates the synergistic co-evolution of the Proposer and Critic.  

\subsection{Propose-then-Critic Paradigm}

Formally, we define the GUI grounding task as locating a target coordinate $p^* \in \mathbb{R}^2$ on a screen $\mathcal{I}$ given an instruction $\mathcal{T}$. We employ a single MLLM, parameterized by $\theta$, which alternates between two distinct roles \textbf{Proposer} and \textbf{Critic} guided by task-specific prompts.

\textbf{The Proposer (Generation).} Given the raw image $\mathcal{I}$ and instruction $\mathcal{T}$, the model operates in generation mode to generate a set of $K$ candidate coordinates $\mathcal{P} = \{p_1, \dots, p_K\}$ in a single inference pass:
\begin{equation}
    \mathcal{P} \sim \pi_{\theta}(\cdot | \mathcal{I}, \mathcal{T}, prompt_{gen})
\end{equation}

\textbf{Visualization (The Bridge).} We employ a visualization function $\mathcal{V}$ to render the candidate set $\mathcal{P}$ onto the screen, generating a "marked image" $\mathcal{I}_{vis} = \mathcal{V}(\mathcal{I}, \mathcal{P})$, where each candidate is annotated with a distinct visual marker (e.g., a numeric tag).

\textbf{The Critic (Discrimination).}  The generation history $\mathcal{H}_{gen} = \{\mathcal{I}, \mathcal{T}, prompt_{gen}, \mathcal{P}\}$. The Critic outputs a ranked sequence of indices $\rho$:
\begin{equation}
    \rho \sim \pi_{\theta}(\cdot | \mathcal{H}_{gen}, \mathcal{I}_{vis}, prompt_{crit})
\end{equation}

where $\rho = (k_1, \dots, k_m)$ is a permutation of the candidate indices. The final prediction is the top-ranked candidate $\hat{p} = p_{k_1}$. This setup ensures the critic evaluates candidates with full awareness of the initial intent and generation context.

\subsection{Proposer and Critic Reward Design}
CoPC formulates accuracy and spatial coverage reward signals for the Proposer to model the precision and diversity of the generation phase. Subsequently, for the Critic phase, it employs Top-1 accuracy and ranking quality as the two distinct reward signals.

\textbf{Proposer Accuracy Reward: }
We quantify the quality of a single point $p_i$ by mapping its Euclidean distance from the ground truth center $p_{gt}$ to the $(0, 1]$ interval using a Gaussian kernel:
\begin{equation}
    s_i = \exp\left(-\frac{\|p_i - p_{gt}\|^2}{2\sigma^2}\right) 
\end{equation}

To aggregate the quality of $K$ samples, we utilize Softmax aggregation to prioritize the optimal candidate, effectively mitigating the impact of suboptimal samples and preventing average-based reward dilution from suppressing generative diversity. Additionally, a binary Hit Reward is added to incentivize falling within the ground truth bounding box $B_{gt}$. The accuracy reward is formulated as:
\begin{equation}
R_{acc} = \sum_{i=1}^{K} \text{Softmax}(s_i) \cdot s_i + \frac{1}{K}\sum_{i=1}^{K} r_{hit, i}
\end{equation}

where $r_{hit, i}$ denotes the hit reward for candidate $p_i$, which equals a defined constant if $p_i$ falls within the ground truth bounding box, and 0 otherwise.

\textbf{Proposer Coverage Reward: }
To facilitate effective Test-time Search, we encourage spatial diversity by quantifying the geometric area covered by the candidate set. We compute the covariance matrix $\Sigma$ of the generated coordinates and map its determinant to a normalized range $[0, 1]$ using a hyperbolic tangent function:
\begin{equation}
    R_{cov} = \tanh \left( \sqrt{| \det(\Sigma) | + \epsilon} \right)
\end{equation}

where $\epsilon$ ensures numerical stability. This formulation prevents unbounded reward growth while incentivizing sufficient exploration.


\textbf{Critic Top-1 Selection Reward: }
To explicitly capture the final prediction accuracy, we formulate the Critic Top-1 Selection Reward. Let $p_{best}$ be the point selected by the Critic as the top candidate. The reward is defined as the selected point's quality:
\begin{equation}
    R_{top1} = s_{best}
\end{equation}

\textbf{Critic Ranking Reward: }
To cultivate fine-grained discrimination, we employ Normalized Discounted Cumulative Gain (NDCG), which imposes heavier penalties on ranking errors at the top of the list. We define the relevance score $s_i$ for a candidate $p_i$ using a Gaussian kernel based on its distance to the ground truth. The reward is calculated as the ratio of the Discounted Cumulative Gain (DCG) to the Ideal DCG (IDCG):
\begin{equation}
    R_{ndcg} = \frac{\text{DCG}_K}{\text{IDCG}_K} = \frac{\sum_{i=1}^{K} s_{\pi_i} / \log_2(i + 1)}{\sum_{i=1}^{K} s_{(i)} / \log_2(i + 1)} 
\end{equation}

where $\pi_i$ denotes the candidate index predicted at rank $i$, and $s_{(i)}$ represents the relevance scores sorted in descending order ($s_{(1)} \ge \dots \ge s_{(K)}$). This ensures the Critic maximizes the alignment of the predicted ranking with the  ground truth.

\textbf{Maturity-Aware Co-Evolution: }
To balance the Proposer's accuracy and diversity while steadily enhancing final prediction performance, we design a progressive dynamic weighting mechanism based on the model's training maturity. This approach effectively mitigates optimization conflicts arising from simultaneously maximizing all rewards from scratch, such as the model prioritizing diversity over accuracy before learning to localize targets.

We define the maturity of the Proposer ($C_P$) and the Critic ($C_J$) using the Exponential Moving Average of their respective performance metrics:
\begin{equation}
    \begin{aligned} C_P^{(t)} &= (1-\alpha)C_P^{(t-1)} + \alpha R_{acc}^{(t)} \\ C_J^{(t)} &= (1-\alpha)C_J^{(t-1)} + \alpha R_{ndcg}^{(t)} \end{aligned}
\end{equation}

where $\alpha$ is the momentum coefficient. The final rewards are dynamically weighted as follows:
\begin{equation}
    \begin{aligned} \mathcal{R}_{Proposer} &= R_{acc} +  C_J \cdot R_{cov} \\ \mathcal{R}_{Critic} &= R_{top1} + C_P \cdot R_{ndcg} \end{aligned}
\end{equation}

\subsection{Why Propose-then-Critic? A Mechanism Analysis}

The core mechanism of COPC utilizes maturity-aware dynamic weighted rewards to simultaneously enhance the model's generative and discriminative capabilities. In this section, we analyze the effectiveness of COPC training methodology.

\textbf{Proposer Evolves Critic: Progressive Hard Negative Mining.}
Driven by the maturity-aware mechanism, the Proposer constructs a dynamic curriculum for the Critic. As Proposer's maturity increases, it evolves from making random errors to generating high-quality candidates that cluster densely around valid UI elements. These progressively harder negatives force the Critic to upgrade its capability from simple noise filtering to fine-grained visual discrimination, ensuring it can distinguish the precise target from semantic distractors.

\textbf{Critic Unlocks Proposer: Adaptive Exploration via Maturity.}
The Critic’s maturation is the catalyst for unlocking the Proposer’s latent potential. In our co-evolutionary strategy, the system dynamically upweights the coverage reward only as the Critic’s reliability improves. This reliable safety net allows the Proposer to shift from conservative regression to expansive sampling. By mitigating the risks of spatial variance, the mature Critic empowers the Proposer to maximize recall, transforming pixel-perfect localization into a manageable high-coverage search.

%% file: tex/experiment.tex
%
\input{tables/final_result_compare}
\subsection{Experiment Settings}
\subsubsection{Datasets.}
We evaluate our method on a total of \textbf{6 benchmarks}. Specifically, we use the training
 sets from Widget Caption\cite{widgetcaption}, OmniAct\cite{omniact}, GUICourse\cite{guicourse}, ShowUI\cite{showui}, RICO-SCA\cite{rico-sca}, OS-ATLAS\cite{atlas} to train our models. Our benchmarks includes MMBench-GUI\cite{mmbench-gui}, ScreenSpot-Pro \cite{screenspotpro}, UI-Vision \cite{uivision}, ScreenSpot-v2 \cite{atlas}, UI-I2E-Bench\cite{ui-e2i}, OSWorld-G\cite{osworldg}. Detail in Appendix C.

 \subsubsection{Baselines.}
To comprehensively evaluate the effectiveness of the COPC method, we benchmark it against three categories of baseline methods:
 \textbf{Direct Prompting on Base Model};
 \textbf{SFT Based Methods}: standard SFT on the dataset, CoT SFT on the dataset, Two Stage SFT on the dataset ;
 \textbf{RL Based Methods}: DPO\cite{dpo}, GRPO\cite{guo2025deepseek}.
 We also selected a range of both open-source and Closed-Source General and GUI-Specific MLLMs as  baselines: Claude 3.7 Sonnet\cite{anthropic2024claude37}, GPT-4o\cite{hurst2024gpt4o}, Claude Computer Use\cite{anthropic2024computeruse}, SEED-1.5-VL\cite{guo2025seed1}, Qwen2.5-VL-3B, Qwen2.5-VL-7B\cite{Qwen2.5-VL}, Qwen3-VL-2B, and Qwen3-VL-4B, Qwen3-VL-8B\cite{Qwen3-VL}, Aguvis-7B\cite{aguvis}, OS-Atlas-Base-7B, UI-TARS-1.5-7B\cite{uitars}, UGround-V1-7B\cite{uground}, GUI-G2-7B\cite{guig2}, SE-GUI-3B, SE-GUI-7B\cite{segui}, GUI-Spotlight\cite{guispotlight}, GUI-Cursor\cite{guicursor}. Detail in Appendix B.
 
\subsubsection{Metrics.}

We employ two metrics to decouple the evaluation of generation coverage and discrimination precision:
\textbf{Oracle@K:} Evaluates the Proposer's recall capability. It calculates the percentage of samples where at least one of the generated candidates ($N \le K$) falls within the ground-truth bounding box.
\textbf{Top-1 Accuracy:} Evaluates the end-to-end system performance. It calculates the percentage of samples where the highest-ranked candidate selected by the Critic falls within the ground-truth bounding box.



\subsection{Comparison with Different Models}
To verify effectiveness, we benchmark against Closed-Source, General Open-Source, and GUI-specific models (Table \ref{tab:gui_comparison_model}). We report direct inference performance based on our reproduction for general models and official results for GUI-specific baselines. Our performance denotes the Top-1 Accuracy.

\textbf{Obs.\ding{182}} \textbf{Our method consistently achieves state-of-the-art performance across varying model scales.} Our method consistently outperforms baselines across varying scales. Notably, on OSWorld-G, Ours (UI-Tars1.5) achieves 48.5\% , surpassing the base model (46.8
\%). Furthermore, Ours (Qwen3-VL-8B) demonstrates robust gains,  validating the efficacy of our COPC framework.

\input{tables/method_compare}
\subsection{Performance of Different Training Methods}
To verify that CoPC can simultaneously enhance the model's candidate generation and visual discrimination accuracy, we conducted an extensive comparison on Qwen3-VL-4B against baselines including Direct SFT, Two-Stage SFT, DPO, and GRPO across multiple benchmarks (Table \ref{tab:main_results}).

\textbf{Obs.\ding{183}} \textbf{SFT methods fail to cultivate self-critic capabilities} 
As indicated by the significant performance degradation gap $\Delta$ in Table \ref{tab:main_results}, SFT methods are insufficient for training the model to visually discriminate the correct target from self-generated distractors.

\textbf{Obs.\ding{184}} \textbf{CoPC achieves improvement in both generation and discrimination.} Unlike previous RL methods that often trade off exploration for stability, CoPC outperforms the strong baseline GRPO on both Oracle@5 and Top-1 accuracy across the vast majority of tasks. This proves that our maturity-aware co-evolution strategy effectively enhances the Proposer's spatial coverage while improve the Critic's capability.

\subsection{Comparison with Spatial Consistency Strategies}
\begin{figure}[h]
  \includegraphics[width=0.47\textwidth]{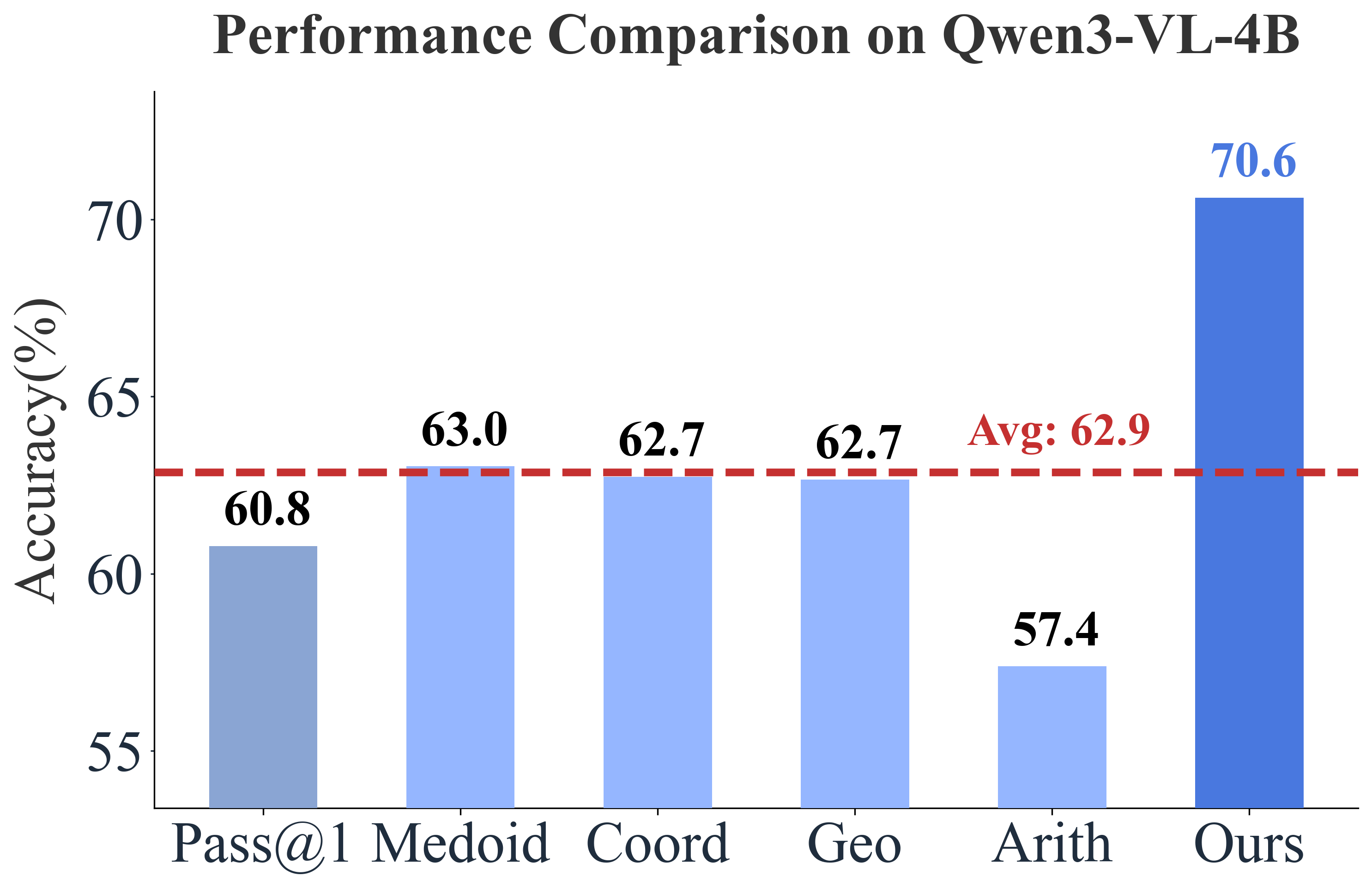}
  \caption{Comparison with Spatial Consistency Strategies.}
  \label{fig:Qwen3vlcopm}
\end{figure}
To validate the necessity of our Visual Critic, we benchmark against four spatial aggregation strategies—Arithmetic Mean, Coordinate-wise Median, Geometric Median, and Medoid—which aggregate answers from 8 independent inference runs. In contrast, our method generates multiple candidates in a single pass followed by one critic turn, significantly reducing token consumption while introducing visual verification. Comparison results on Qwen3-VL-4B are shown in Figure \ref{fig:Qwen3vlcopm}.

\textbf{Obs.\ding{185} Visual Discrimination surpasses geometric consensus with higher efficiency.} 
Despite consuming significantly fewer tokens, our method surpasses all spatial baselines. We observe that Arithmetic Mean often fail predictions into invalid background regions (e.g., between icons). Similarly, Medoid and Median fail on multi-modal distributions: they blindly select the geometric center, whereas our Critic explicitly uses visual cues to distinguish the true target from semantic distractors.

\subsection{Sensitivity Analysis}
\begin{figure}[h]
  \includegraphics[width=0.47\textwidth]{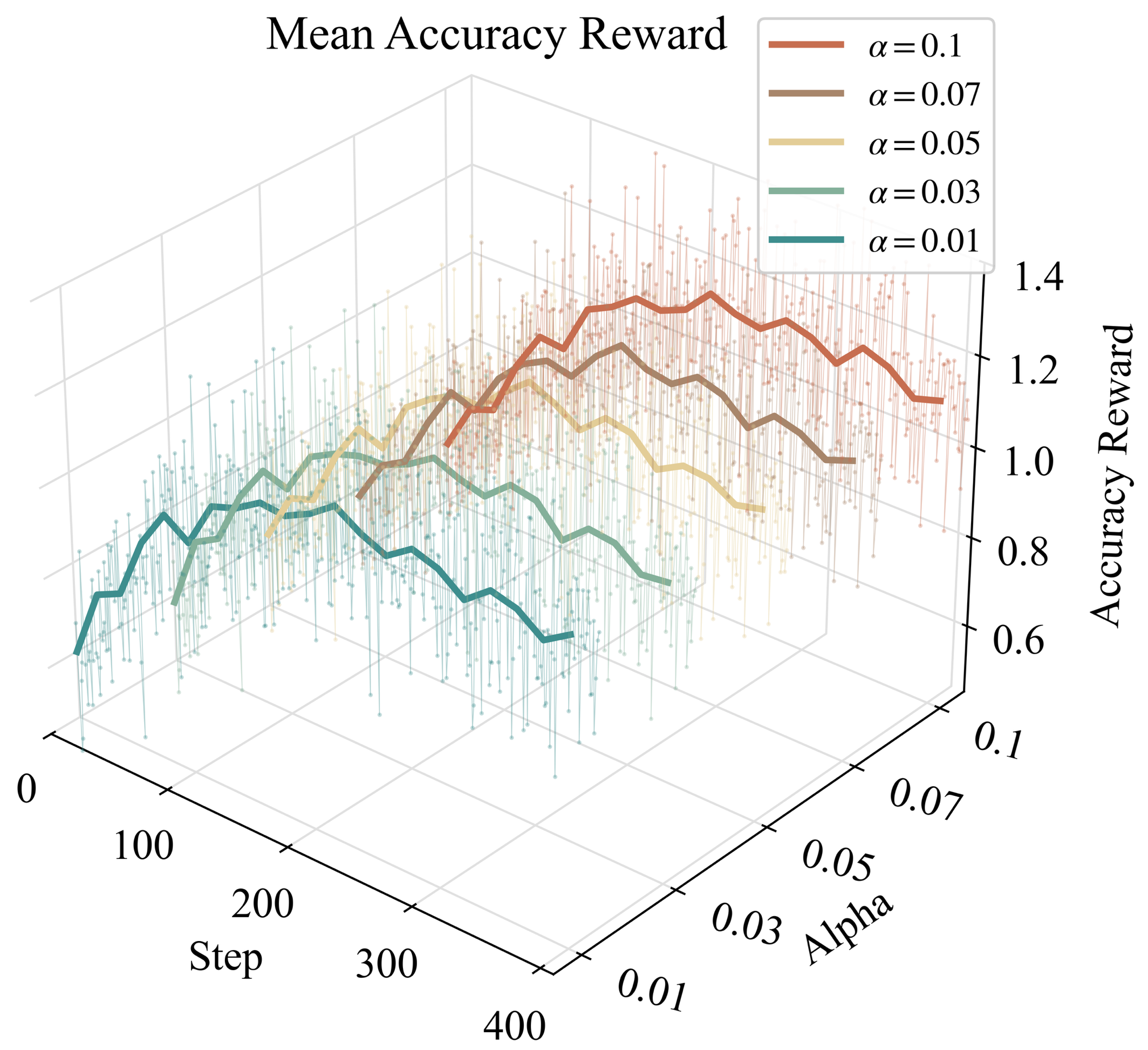}
  \caption{Sensitivity Analysis of Momentum Coefficient $\alpha$ on Qwen3-VL-2B.}
  \label{fig:sensitive}
\end{figure}
To evaluate the stability of our maturity-aware mechanism, we analyze the impact of the momentum coefficient $\alpha$ on Qwen3-VL-2B , as shown in Figure \ref{fig:sensitive}.

\textbf{Obs.\ding{186} Robustness to maturity update rates.}
We observe that Mean Accuracy Reward curves exhibit consistent convergence with minimal fluctuation across a broad range of $\alpha$. This confirms our framework's robustness to momentum settings, ensuring stable optimization without extensive tuning.

\subsection{Performance of CoPC on Different Models}
\begin{figure}[ht]
  \centering
  \begin{minipage}[c]{0.234\textwidth}
    \includegraphics[width=\textwidth]{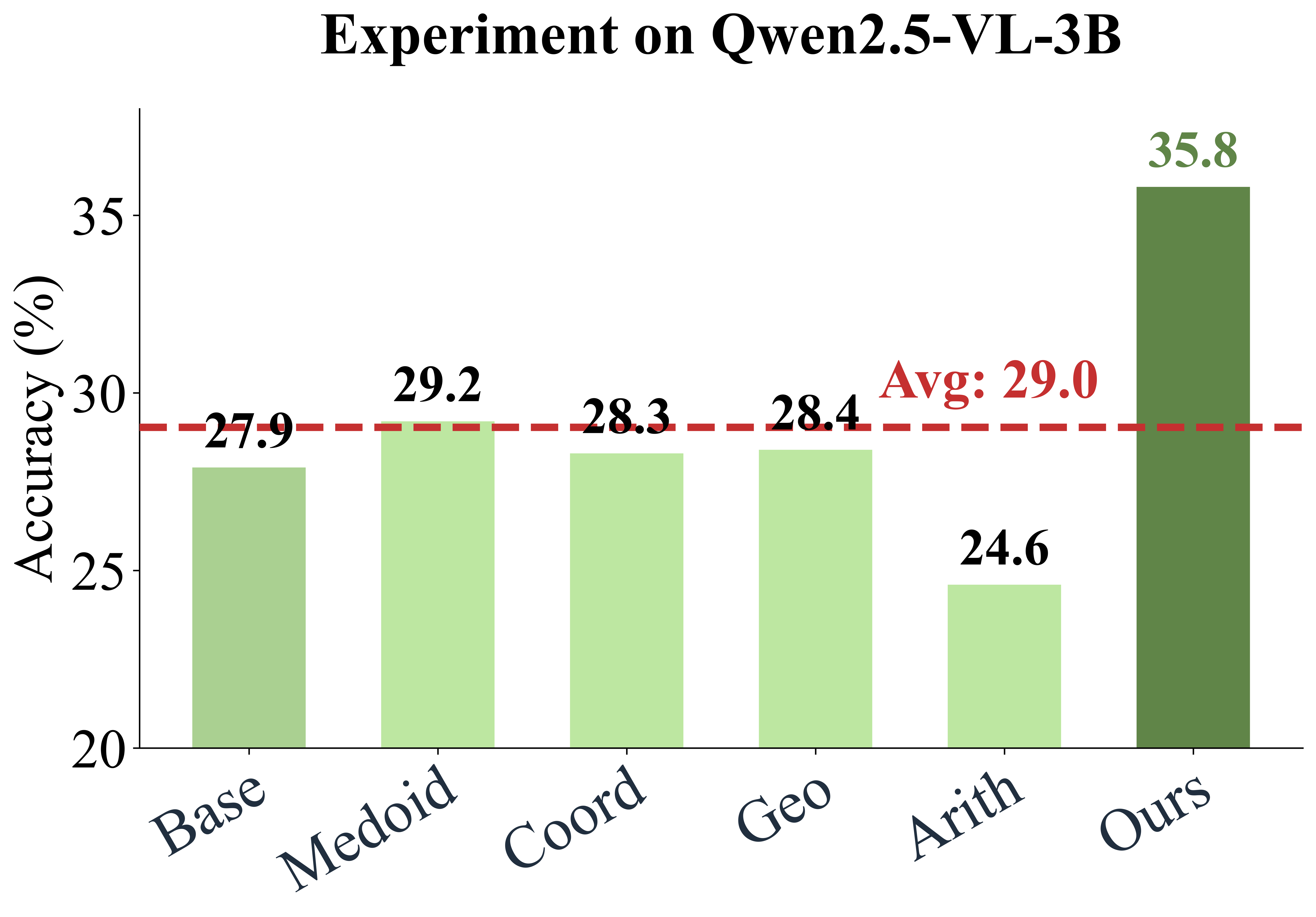}
  \end{minipage}
  \begin{minipage}[c]{0.234\textwidth}
    \includegraphics[width=\textwidth]{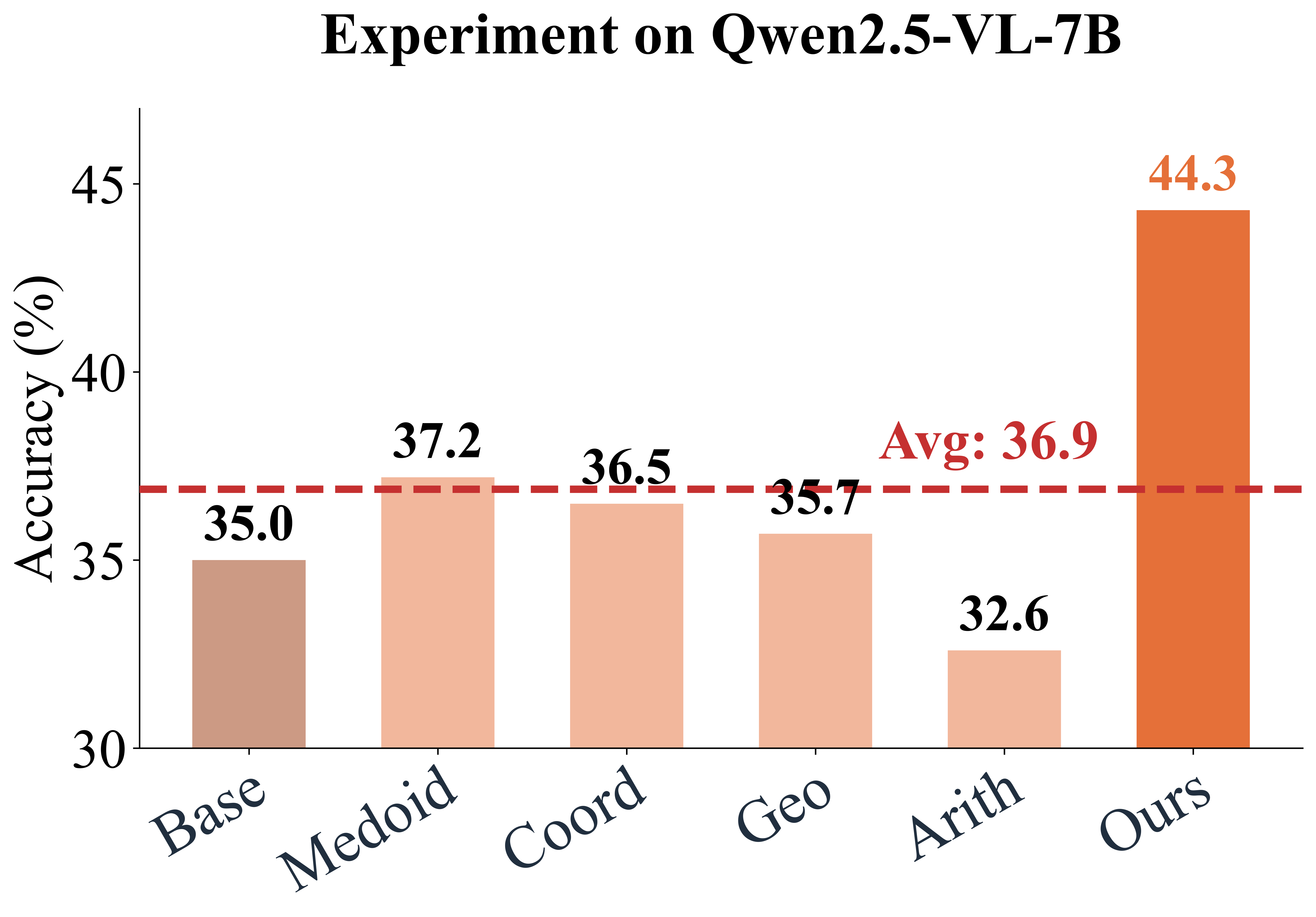}
  \end{minipage}

  \caption{Performance Analysis on Qwen2.5-VL Series.}
  \label{fig_different_model}
\end{figure}

To verify the generality of our framework, we extended our evaluation to the Qwen2.5-VL series (3B and 7B), as shown in Figure \ref{fig_different_model}.

\textbf{Obs.\ding{187} achieves superior training performance
across a variety of models. }Our method consistently outperforms both base models and spatial aggregation strategies regardless of the underlying architecture, demonstrating that the benefits of our co-evolutionary mechanism are transferable across different backbones.
\subsection{Ablation Study}
\input{tables/ablation}
We conduct ablation studies on Qwen3-VL-4B to evaluate the contribution of each component, as shown in Table \ref{tab:ablation_study_delta}.

\textbf{Obs.\ding{188} Maturity Mechanism ensures stable optimization.} Replacing the dynamic maturity schedule with static weights results in the most severe performance drop. This confirms that dynamic weighting is essential to resolve the optimization conflict between accuracy and diversity.

\textbf{Obs.\ding{189} Specific rewards are indispensable for training.} Removing $R_{cov}$ leads to a notable decline in Oracle@5, verifying its necessity for maximizing candidate recall.
Removing $R_{ndcg}$ widens the selection gap, indicating that ranking objectives are crucial for the Critic to pinpoint the correct target from valid candidates

%% file: tables/final_result_compare.tex
\begin{table*}[t!]
\centering
\renewcommand{\arraystretch}{0.8}
\setlength{\tabcolsep}{1.7pt}
\begin{tabular}{@{}l*{8}{S[table-format=2.2]}@{}}
\toprule[1.5pt]

\textbf{Model} & {\textbf{SSpot-Pro}} & {\textbf{SSpot-v2}} & {\textbf{OSWorldG}} & {\textbf{MMB-L2}} & {\textbf{I2E}} & {\textbf{UIV-B}} & {\textbf{UIV-F}} & {\textbf{UIV-S}} \\
\midrule
\multicolumn{9}{c}{\textit{Closed-Source Models}} \\
\midrule
GPT-4o & 0.8 & 20.1 & {\textendash} & 2.87 & {\textendash} & 1.6 & 1.5 & 1.0 \\
Claude-3.7 Sonnet & 27.7 & 87.6 & {\textendash} & 4.66 & {\textendash} & 9.5 & 7.7 & 7.6 \\
Claude Computer Use & 17.1 & {\textendash} & {\textendash} & {\textendash} & {\textendash} & {\textendash} & {\textendash} & {\textendash} \\
Seed-1.5-VL & 60.9 & 95.2 & 62.9 & {\textendash} & {\textendash} & {\textendash} & {\textendash} & {\textendash} \\
Operator & 36.6 & 70.5 & 40.6 & {\textendash} & {\textendash} & {\textendash} & {\textendash} & {\textendash} \\
\midrule
\multicolumn{9}{c}{\textit{General Open-source Models}} \\
\midrule
Qwen2.5-VL-3B & 16.1 & 80.9 & 27.3 & 21.2 & 21.7 & 0.6 & 0.2 & 0.3 \\
Qwen2.5-VL-7B & 26.8 & 88.8 & 31.4 & 33.8 & 28.4 & 1.2 & 0.8 & 0.5 \\
Qwen3-VL-2B & 16.7 & 70.3 & 23.8 & 48.1 & 44.8 & 11.7 & 9.3 & 2.1 \\
Qwen3-VL-4B & 46.4 & 88.7 & 55.3 & 77.4 & 72.1 & 27.2 & 26.5 & 8.8 \\
Qwen3-VL-8B & 50.1 & 91.2 & 54.7 & 80.1 & 76.5 & 35.1 & 30.6 & 10.8 \\
\midrule
\multicolumn{9}{c}{\textit{GUI-specific Models}} \\
\midrule
Aguvis-7B & {\textendash} & {\textendash} & 38.7 & {\textendash} & {\textendash} & 17.8 & 18.3 & 5.1 \\
OS-Atlas-Base-7B & 18.9 & 85.1 & {\textendash} & 41.4 & {\textendash} & 12.2 & 11.2 & 3.7 \\
UI-TARS-1.5-7B & 42.0 & 91.6 & 46.8 & 64.3 & 66.1 & 22.9 & 25.1 & 6.6 \\
UGround-V1-7B & 31.1 & {\textendash} & {\textendash} & 65.6 & {\textendash} & 15.4 & 17.1 & 6.3 \\
GUI-G2-7B & 47.5 & 93.3 & {\textendash} & {\textendash} & {\textendash} & {\textendash} & {\textendash} & {\textendash} \\
GUI-Spotlight & 52.8 & {\textendash} & 62.7 & {\textendash} & {\textendash} & 32.1 & 30.2 & 9.1 \\
GUI-Cursor & 56.5 & 93.9 & {\textendash} & {\textendash} & {\textendash} & {\textendash} & {\textendash} & {\textendash} \\
\midrule
\rowcolor{gray!15}
\textbf{Ours (UI-Tars1.5)} & \textbf{47.3} & \textbf{90.2} & \textbf{48.5} & \textbf{68.9} & \textbf{69.3} & \textbf{24.0} & \textbf{22.9} & \textbf{7.4} \\
\rowcolor{gray!15}
\textbf{Ours (Qwen3-VL-8B)} & \textbf{58.7} & \textbf{91.3} & \textbf{59.6} & \textbf{78.4} & \textbf{81.8} & \textbf{38.2} & \textbf{34.5} & \textbf{14.1} \\
\bottomrule[1.5pt]
\end{tabular}
\caption{Comparison of our model against various Closed-Source, General Open-Source, and GUI-specific MLLMs on 6 GUI-related benchmarks.}
\label{tab:gui_comparison_model}
\end{table*}

%% file: tables/method_compare.tex
\begin{table*}[t]
\centering

\setlength{\tabcolsep}{1.7pt} 
\renewcommand{\arraystretch}{0.8}
\begin{tabular}{l
                S[table-format=2.1] S[table-format=2.1] r
                S[table-format=2.1] S[table-format=2.1] r
                S[table-format=2.1] S[table-format=2.1] r
                S[table-format=2.1] S[table-format=2.1] r
               }
\toprule
\multirow{2}{*}{\textbf{Method}} & 
\multicolumn{3}{c}{\textbf{ScreenSpot-Pro}} & \multicolumn{3}{c}{\textbf{ScreenSpot-v2}} & \multicolumn{3}{c}{\textbf{OSWorld-G}} & \multicolumn{3}{c}{\textbf{MMBench-GUI}} \\
\cmidrule(lr){2-4} \cmidrule(lr){5-7} \cmidrule(lr){8-10} \cmidrule(lr){11-13}

& {Ora.@5} & {Top-1} & {$\Delta$} & {Ora.@5} & {Top-1} & {$\Delta$} & {Ora.@5} & {Top-1} & {$\Delta$} & {Ora.@5} & {Top-1} & {$\Delta$} \\
\midrule

Base            & 41.2 & 29.3 & \textcolor{myred}{-11.9} & 91.3 & 79.3 & \textcolor{myred}{-12.0} & 51.1 & 32.8 & \textcolor{myred}{-18.3} & 77.9 & 58.3 & \textcolor{myred}{-19.6} \\
Direct SFT      & 32.4 & 23.6 & \textcolor{myred}{-8.8}  & 84.7 & 65.1 & \textcolor{myred}{-19.6} & 42.6 & 25.7 & \textcolor{myred}{-16.9} & 68.9 & 47.2 & \textcolor{myred}{-21.7} \\
Two-Stage SFT   & 43.7 & 35.8 & \textcolor{myred}{-7.9}  & 90.6 & 85.4 & \textcolor{myred}{-5.2}  & 52.9 & 47.5 & \textcolor{myred}{-5.4}  & 74.8 & 64.3 & \textcolor{myred}{-10.5} \\
DPO             & 42.5 & 34.2 & \textcolor{myred}{-8.3}  & 90.3 & 88.6 & \textcolor{myred}{-1.7}  & 52.3 & 48.2 & \textcolor{myred}{-4.1}  & 72.4 & 65.9 & \textcolor{myred}{-6.5} \\
GRPO            & 58.4 & 53.4 & \textcolor{myred}{-5.0}  & \textbf{93.1} & 92.5 & \textcolor{myred}{-0.6}  & 61.2 & 57.3 & \textcolor{myred}{-3.9}  & 82.9 & 75.7 & \textcolor{myred}{-7.2} \\
\midrule

\rowcolor{gray!20}
\textbf{Ours}   & \textbf{59.8} & \textbf{56.4} & \textbf{\textcolor{myred}{-3.4}} & \textbf{93.1} & \textbf{92.6} & \textbf{\textcolor{myred}{-0.5}} & \textbf{63.2} & \textbf{59.4} & \textbf{\textcolor{myred}{-3.8}} & \textbf{84.5} & \textbf{79.3} & \textbf{\textcolor{myred}{-5.2}} \\
\bottomrule
\end{tabular}
\end{table*}

\begin{table*}[t!]
\centering
\setlength{\tabcolsep}{1.7pt} 
\renewcommand{\arraystretch}{0.8}
\begin{tabular}{l
                S[table-format=2.1] S[table-format=2.1] r
                S[table-format=2.1] S[table-format=2.1] r
                S[table-format=2.1] S[table-format=2.1] r
                S[table-format=2.1] S[table-format=2.1] r
               }
\toprule
\multirow{2}{*}{\textbf{Method}} & 
\multicolumn{3}{c}{\textbf{I2E-bench}} & \multicolumn{3}{c}{\textbf{UI-Vision-Basic}} & \multicolumn{3}{c}{\textbf{UI-Vision-Func}} & \multicolumn{3}{c}{\textbf{UI-Vision-Spatial}} \\
\cmidrule(lr){2-4} \cmidrule(lr){5-7} \cmidrule(lr){8-10} \cmidrule(lr){11-13}

& {Ora.@5} & {Top-1} & {$\Delta$} & {Ora.@5} & {Top-1} & {$\Delta$} & {Ora.@5} & {Top-1} & {$\Delta$} & {Ora.@5} & {Top-1} & {$\Delta$} \\
\midrule

Base            & 74.2 & 55.0 & \textcolor{myred}{-19.2} & 25.8 & 16.4 & \textcolor{myred}{-9.4} & 27.5 & 16.6 & \textcolor{myred}{-10.9} & 17.6 & 6.1  & \textcolor{myred}{-11.5} \\
Direct SFT      & 63.2 & 49.3 & \textcolor{myred}{-13.9} & 18.6 & 12.1 & \textcolor{myred}{-6.5} & 22.0 & 13.4 & \textcolor{myred}{-8.6}  & 8.2  & 2.3  & \textcolor{myred}{-5.9} \\
Two-Stage SFT   & 73.4 & 65.1 & \textcolor{myred}{-8.3}  & 26.3 & 20.2 & \textcolor{myred}{-6.1} & 26.4 & 18.3 & \textcolor{myred}{-8.1}  & 17.9 & 13.4 & \textcolor{myred}{-4.5} \\
DPO             & 76.6 & 62.3 & \textcolor{myred}{-14.3} & 22.1 & 15.9 & \textcolor{myred}{-6.2} & 25.8 & 18.0 & \textcolor{myred}{-7.8}  & 16.4 & 12.3 & \textcolor{myred}{-4.1} \\
GRPO            & 78.4 & 72.6 & \textcolor{myred}{-5.8}  & 32.9 & 25.0 & \textcolor{myred}{-7.9} & 34.2 & 26.0 & \textcolor{myred}{-8.2}  & 17.3 & 12.8 & \textcolor{myred}{-4.5} \\
\midrule

\rowcolor{gray!20}
\textbf{Ours}   & \textbf{80.6} & \textbf{76.0} & \textbf{\textcolor{myred}{-4.6}} & \textbf{36.8} & \textbf{30.5} & \textbf{\textcolor{myred}{-6.3}} & \textbf{34.6} & \textbf{27.8} & \textbf{\textcolor{myred}{-6.8}} & \textbf{19.4} & \textbf{15.5} & \textbf{\textcolor{myred}{-3.9}} \\
\bottomrule
\end{tabular}
\caption{
    Performance comparison across 6 GUI Grounding benchmarks. We report Oracle@5 (generator upper bound), Top-1 (final selection), and the performance gap ($\Delta$). Our method consistently achieves the best accuracy with the smallest gap (closest to 0), demonstrating effective visual discrimination.
}
\label{tab:main_results}
\end{table*}

%% file: tables/ablation.tex
\begin{table}[h]
\centering
\renewcommand{\arraystretch}{0.8}
\setlength{\tabcolsep}{1.6pt}
\begin{tabular}{l S[table-format=2.1] S[table-format=2.1] r}
\toprule
\textbf{Method} & {\textbf{Ora.@5}} & {\textbf{Top-1}} & {\textbf{$\Delta$}} \\
\midrule
\rowcolor{gray!15}
\textbf{Ours (Full Model)}      & \textbf{70.0} & \textbf{67.3} & \textbf{-2.7} \\
\midrule
\multicolumn{4}{l}{\textit{w/o component:}} \\
--- Maturity (Static Weights)   & 62.6          & 57.5 & \textcolor{red}{-5.1} \\
--- Coverage Reward             & 66.7          & 63.2 & \textcolor{red}{-3.5} \\
--- Ranking Reward              & 69.3          & 64.9 & \textcolor{red}{-4.4} \\
\bottomrule
\end{tabular}
\caption{Ablation study on Qwen3-VL-4B.}
\label{tab:ablation_study_delta}
\end{table}

%% file: tex/conclusion.tex
In this paper, we introduce the Co-evolutionary Framework of Proposer and Visual Critic (COPC), shifting GUI grounding to a Visual Perception Ranking paradigm. Utilizing a maturity-aware co-evolutionary RL strategy, we synergistically optimize the Proposer and Critic to unlock latent potential. Experiments across 6 benchmarks confirm COPC outperforms state-of-the-art models. By bridging the semantic-visual gap, our method advances robust autonomous GUI agents.

%% file: tex/limitations.tex
While CoPC demonstrates significant improvements in GUI grounding precision, there remain marginal areas for refinement. First, as a two-stage framework, although we have substantially reduced computational overhead compared to standard consistency strategies, the inference latency is inherently higher than single-step regression models, presenting a trade-off for strictly real-time applications. Second, in extremely dense interface layouts, the visual markers rendered for critic verification might occasionally introduce slight occlusion to local semantic details of tiny elements. Future work will explore more lightweight architectures and adaptive visual prompting techniques to address these aspects.

%% file: tex/acknowledgements.tex
Supported by National Natural Science Foundation of China (No. 62402429, U24A20326, 62441236)
Supported by the Key Research and Development Program of Zhejiang Province(No. 2025C01026)
Ningbo Yongjiang Talent Introduction Programme (2023A-397-G)
Young Elite Scientists Sponsorship Program by CAST (2024QNRC001)
The author gratefully acknowledges the support of Zhejiang University Education Foundation Qizhen Scholar Foundation.
Supported by Tencent Rhino-Bird Joint Research Program

%% file: tex/appendix.tex

\label{sec:appendix}
\section{EXPERIMENTAL SETUP}

In this section, we provide a detailed description of the experimental configuration, including a comprehensive explanation of training dataset, dataset filtering strategy, base algorithm, training hyperparameter and prompt design.

\subsection{Training Dataset}
We utilized the training sets from Widget Caption, OmniAct, GUICourse, ShowUI, RICO-SCA, and OS-ATLAS to train our model. These datasets cover a variety of GUI environments including mobile, web, and desktop platforms. We use the dataset strictly for its intended purpose of evaluating algorithms. We state that we used the pre-processed version of datasets, where the original authors have already removed PII and offensive content.

\textbf{Widget Captioning} dataset is a large-scale crowdsourced dataset for automatic UI element captioning. Based on the extended, crawler-supplemented RICO dataset and filtered, it includes 21,750 unique UI screens (6,470 apps), 61,285 caption-missing elements, and 162,859 human annotations. Covering 10 categories (e.g., image buttons, switches), its descriptions are mostly 2-3 words (avg. 2.72) with a "predicate + object" structure, plus multimodal data (UI screenshots, view hierarchies) for model training(CC BY 4.0).

\textbf{OmniACT} dataset is the first multimodal dataset for desktop and web apps, with 9,800 pairs of screenshots, natural language instructions, and PyAutoGUI executable scripts. Supporting cross-platform use (macOS, Windows, Linux, web) and 6 task categories, it enables training and evaluating generalist autonomous agents integrating language and visual UI understanding.

\textbf{GUICourse} is a dataset suite for training vision-based GUI agents from VLMs, including GUIEnv (10M+0.7M samples to boost OCR/grounding), GUIAct (67k single-step, 5.6k multi-step web, 9.1k smartphone samples for navigation), and GUIChat (44k single-turn, 6k multi-turn dialogues for interaction), addressing VLMs’ GUI limitations via foundational ability and knowledge enhancement(Creative Commons Attribution 4.0 International License).

\textbf{ShowUI} introduces a small-scale yet high-quality instruction-following dataset tailored for training GUI visual agents, addressing critical data challenges in GUI perception and interaction. Curated through meticulous data analysis and a rebalanced sampling strategy, this dataset encompasses 256K task instances (screenshots for grounding, tasks for navigation) with 2.7M element annotations, spanning diverse devices including web, mobile, and desktop platforms(Apache-2.0 license).

\textbf{RICO-SCA}, a synthetic grounding dataset, includes 295k single-step commands for 178k UI objects across 25k+ Android screens (from Rico). ANDROIDHOWTO, for action phrase extraction, has 32k data points from 9k+ web how-to guides with annotated operation/object/argument spans. PIXELHELP, for full-task evaluation, provides 187 multi-step Pixel phone instructions paired with human-executed action-screen sequences(apache-2.0).

\textbf{OS-Atlas} is the largest open-source cross-platform GUI grounding corpus to date, developed to address the critical data scarcity in training generalist GUI agents. It comprises over 2.3 million distinct screenshots and more than 13.58 million GUI elements, spanning five major platforms—Windows, macOS, Linux, Android, and the web—filling the long-standing gap of desktop GUI grounding data in existing datasets(apache-2.0). 

\subsection{Hardware}
All experiments are performed on a CentOS Linux 7 server. The hardware specifications consist of 240GB of RAM, a 16-core Intel Xeon CPU, and eight NVIDIA H20 GPUs, each having 96GB of memory. 

\subsection{Data Filtering Strategy}
To ensure the stability of the co-evolutionary training and maximize the sample efficiency of our reinforcement learning framework, we implement a rigorous, hierarchical data filtering protocol. This process begins with static quality verification, designed to purify the raw instruction-screenshot pairs by eliminating noise and ambiguity. Specifically, we first apply heuristic rules to filter out samples with significant annotation errors, such as invalid coordinates that fall outside the viewport or bounding boxes with extreme scales (e.g., covering less than $0.1\%$ or more than $90\%$ of the screen area), which typically indicate non-interactive background elements or layout noise. Following this, to address the challenge of semantic ambiguity—where an instruction might plausibly refer to multiple elements or none at all—we employ a strong proprietary Vision-Language Model (e.g., Qwen3-VL-235B-A22B) as a zero-shot verifier. The model is presented with the screenshot, the annotated bounding box, and the instruction, and is tasked with verifying their alignment. Samples where the VLM detects a semantic mismatch or ambiguity are excluded, ensuring that our reward signals are grounded in high-quality supervision.

Upon establishing a clean base dataset, we proceed to difficulty-aware pre-selection to isolate the most valuable samples for the RL phase. We define a training subset $\mathcal{D}^{\prime} \subset \mathcal{D}$ by evaluating the initial policy $\pi_{\theta_{0}}$ against the data. For each query $q$, we sample a set of $N$ trajectories. Queries exhibiting "Uniform Success" (where all $N$ trajectories are correct) are discarded as trivial, as the model has already mastered these patterns. Conversely, queries resulting in "Uniform Failure" (where all $N$ trajectories fail) are also removed. Such persistent failure typically stems from overly complex reasoning chains beyond the model's current capacity or residual label noise, which can destabilize the reward mechanism. By filtering these extremes, we focus the training on the "learning frontier"—queries yielding a mixed distribution of outcomes—where the policy's self-evaluation and exploration mechanisms are most actively challenged.

Finally, we enforce online optimization filtering during the on-policy training updates to maintain the fidelity of the gradient estimation. Within the Grouped Reward Policy Optimization (GRPO) framework, policy updates rely on the normalized advantage of sampled trajectories. We identify and discard "Zero-Advantage" mini-batches where all generated trajectories within a group receive identical rewards. In such cases, the advantage estimates vanish, rendering the update non-informative and driven solely by the KL-divergence penalty. Additionally, we implement truncation filtering to exclude trajectories terminated prematurely by the maximum sequence length. This prevents the model from being unfairly penalized for verbose but potentially correct reasoning chains, thereby preserving the accuracy of the critic's ranking reward.

\subsection{Base Algorithm}
To facilitate the stable co-evolution of the Proposer and Critic, we employ Grouped Reward Policy Optimization (GRPO). Unlike standard Proximal Policy Optimization (PPO) which relies on a separate value function critic for advantage estimation, GRPO optimizes the policy by normalizing rewards within a group of outputs generated from the same input. This is particularly advantageous for our framework as it reduces the memory overhead of maintaining value networks for both Proposer and Critic roles while effectively handling the high variance inherent in spatial coordinate sampling.

\textbf{Group-wise Advantage Estimation.}
For each training step, given a batch of input instructions and GUI images $\{(T, I)\}$, we sample a group of $G$ outputs $\{o_1, o_2, \dots, o_G\}$ from the old policy $\pi_{\theta_{old}}$. In the Proposer phase, these outputs correspond to candidate coordinate sets; in the Critic phase, they correspond to ranking permutations.
Each output $o_i$ receives a task-specific reward $r_i$ (as defined in Eq. 4-7). To mitigate the baseline variance across different GUI layouts, we compute the advantage $A_i$ by normalizing the reward against the group statistics:
\begin{equation}
    A_i = \frac{r_i - \text{mean}(\{r_1, \dots, r_G\})}{\text{std}(\{r_1, \dots, r_G\}) + \epsilon}
\end{equation}

where $\epsilon$ is a small constant for numerical stability. This group-wise relative comparison encourages the model to distinguish superior candidates from mediocre ones within the specific context of the current screen, rather than learning absolute reward values.

\textbf{Objective Function.}
The training objective maximizes the surrogate loss while constraining the policy update via KL-divergence regularization to prevent mode collapse. The overall objective function is formulated as follows:
\begin{equation}
\begin{split}
\mathcal{J}(\theta) = \mathbb{E}_{q \sim \mathcal{D}, \{o_i\}} \Bigg[ \frac{1}{G} \sum_{i=1}^G \bigg( & \min \bigg( \frac{\pi_{\theta}(o_i|q)}{\pi_{\theta_{old}}(o_i|q)} A_i, \\
& \mkern-120mu \text{clip}\left(\frac{\pi_{\theta}(o_i|q)}{\pi_{\theta_{old}}(o_i|q)}, 1-\varepsilon, 1+\varepsilon\right) A_i \bigg) \\
& \mkern-120mu - \beta \mathbb{D}_{KL}\big(\pi_{\theta} || \pi_{\text{ref}}\big) \bigg) \Bigg]
\end{split}
\end{equation}

Here, $q$ represents the multimodal input context $(T, I)$. The first term is the clipped surrogate objective standard in PPO, where $\varepsilon$ controls the trust region size. The second term, $\mathbb{D}_{KL}(\pi_{\theta} || \pi_{\text{ref}})$, serves as a token-level regularization penalty, ensuring the co-evolving policy does not deviate excessively from the initial SFT reference policy $\pi_{\text{ref}}$, thereby maintaining the model's fundamental linguistic and visual capabilities during the RL phase.

\subsection{Training Hyperparameter}
In this section, we detail the specific implementation configurations to ensure the reproducibility of our experiments. We adopt Qwen3-VL-8B as the primary backbone for the CoPC framework. The specific hyperparameters governing the Grouped Reward Policy Optimization (GRPO) process and the maturity-aware mechanism—optimized for training stability and convergence—are summarized in Table \ref{tab:hyperparameters_and_training_schedule_large}.
\input{tables/hyperparameters}
\subsection{Prompt Design}
We design two distinct system prompts to drive the propose-then-critic framework. The Proposer prompt instructs the model to employ Chain-of-Thought (CoT) reasoning—analyzing UI layout and user intent—before generating $K$ diverse candidate coordinates $(x, y)$ to maximize target coverage. Conversely, the Critic prompt tasks the model with acting as a visual judge. It receives the screenshot with visualized candidates and strictly ranks them based on a priority hierarchy of target hit accuracy, centrality, and boundary proximity. Both prompts enforce structured JSON output for efficient parsing and interaction.

\textbf{Proposer Prompt:}

\textit{You are a GUI exploration agent acting as a Candidate Generator.
Your goal is to propose a set of potential coordinates that might match the user's instruction. }

\textit{Instructions:}

\textit{First, output your detailed thinking process (100-200 words) to explain:
1. How you interpret the user's core intent (e.g., "click a submit button", "select a dropdown menu");
2. Reasoning about potential target element positions (based on common GUI layout conventions, e.g., "submit buttons are often at the bottom center/right of forms");
3. How you ensure coordinate diversity (e.g., "spreading points across top-left, bottom-center, right-middle areas to cover ambiguous targets");
4. Why you selected these 5 specific coordinates (e.g., "covering likely target regions while avoiding clustering").}

\textit{Then, strictly follow the below rules to generate coordinates:
1. Analyze the UI layout and the user's intent.
2. Generate 5 distinct candidate coordinates $(x, y)$.
3. Do not worry about being perfectly precise with every point. Instead, focus on diversity and coverage. Ensure that at least one of your candidates falls exactly on the target element. 
4. Avoid outputting all points in the same cluster; spread them out slightly to cover different parts of the potential target or ambiguous elements.
  }
\textit{Output STRICTLY in the following JSON block:
```
\{
  "candidates": [
    [x1, y1],
    [x2, y2],
    [x3, y3],
    [x4, y4],
    [x5, y5]
  ]
\}}```
\textit{User Instruction:\{User Instruction\}}

\textbf{Critic Prompt:}

\textit{You are a Visual Judge for GUI navigation.
Input: A screenshot with N candidate positions (IDs 0 to N-1) and a user instruction.}

\textit{Task: Rank ALL candidate IDs (0 to N-1) from most accurate to least accurate based on the criteria below.}

\textit{Ranking Criteria (Strict Priority):
1. Target Hit: Candidates physically ON the correct target element strictly outrank those OFF it.
2. On-Target: Among candidates ON the target, prioritize those closer to the visual center.
3. Off-Target: Among candidates OFF the target, prioritize those closer to the target's boundary.}

\textit{Output Instructions:
1. Analysis Phase: First, explicitly analyze the candidates based on the criteria. Identify which IDs are on the target and which are off.
2. JSON Phase: After the analysis, output the final ranking in a JSON block.}

\textit{Format Example:
[Analysis of candidates...]}

\textit{```json
\{
  "ranked\_ids": [id\_best, id\_second, ..., id\_worst]
\}```}
\section{Baselines}

\subsection{Training Baselines}
To comprehensively evaluate the effectiveness of our proposed framework, we compare it against four distinct training paradigms.

\textbf{Direct Supervised Fine-Tuning (Direct SFT)} represents the standard approach for training GUI agents, where the model is optimized to map the instruction and screenshot directly to the final pixel coordinates. We treat this as a standard regression task using the original dataset triples $(I, T, p_{gt})$ without generating any intermediate reasoning or candidates. The model is trained using the cross-entropy loss to maximize the likelihood of the ground truth coordinates. For the hyperparameters, we utilize the AdamW optimizer with a global batch size of 128 and an initial learning rate of $2 \times 10^{-5}$. We employ a cosine decay scheduler with a warmup ratio of 0.03 over 3 training epochs, setting the weight decay to 0.05 and the maximum sequence length to 6000 to accommodate the visual tokens and textual instructions.

\textbf{Two-Stage Supervised Fine-Tuning (Two-Stage SFT)} is designed to validate the structural advantage of the "Propose-then-Critic" paradigm using purely supervised learning. Since the original dataset lacks intermediate supervision, we construct a distilled dataset using a larger teacher model to generate chain-of-thought reasoning, multiple candidate proposals (Proposer), and the subsequent selection (Critic) for each sample. The student model is then trained to mimic this sequential process in a unified inference session. We maintain the optimization settings similar to Direct SFT, using a learning rate of $2 \times 10^{-5}$ and a batch size of 128 over 3 epochs. However, to handle the increased context length required for the intermediate reasoning steps and candidate lists, we extend the maximum sequence length to 12000.

\textbf{Direct Preference Optimization (DPO)} is introduced to further enhance the model's discrimination capabilities by incorporating negative constraints into the Two-Stage SFT framework. Building upon the supervised checkpoint, we construct preference pairs where trajectories that successfully locate the target (Hit) are labeled as chosen responses $y_w$, and those that fail to hit the target are labeled as rejected $y_l$. The model is optimized to maximize the implicit reward margin between these positive and negative pairs. For this alignment stage, we use a significantly lower learning rate of $5 \times 10^{-7}$ to ensure stability and set the KL penalty coefficient $\beta$ to 0.1. The training is conducted for 1 epoch with a global batch size of 64 and a warmup ratio of 0.1.

\textbf{Group Relative Policy Optimization (GRPO)} serves as a baseline to isolate the impact of our specific reward design and maturity mechanism. While this method utilizes the same two-stage rollout architecture as our proposed CoPC, it is trained solely using basic Accuracy Rewards. Specifically, the Proposer receives a reward only if the ground truth is included in the candidate set. Crucially, the advanced coverage rewards, ranking rewards (NDCG), and the maturity-aware dynamic weighting mechanism are removed. We train this baseline with a learning rate of $1 \times 10^{-6}$, a group size of $G=8$ for advantage estimation, and a KL divergence coefficient of 0.04. The optimization runs for approximately 1 epoch with a batch size of 64 and a maximum sequence length of 12000.

\subsection{Spatial Consistency Baselines}

In Section 5.4, we compare our method against several statistical aggregation strategies to validate the necessity of visual discrimination. Let $\mathcal{P} = \{p_1, p_2, \dots, p_K\}$ denote the set of $K$ candidate coordinates generated by the Proposer for a single instruction, where each $p_i \in \mathbb{R}^2$. The final predicted coordinate $\hat{p}$ is derived using the following formulations:

\textbf{Arithmetic Mean (Mean)}
The Arithmetic Mean calculates the centroid of the candidate set. It is equivalent to minimizing the sum of squared Euclidean distances (L2 loss) to all points. While computationally efficient, it is highly sensitive to outliers.
\begin{equation}
    \hat{p}_{\text{mean}} = \frac{1}{K} \sum_{i=1}^{K} p_i
\end{equation}

\textbf{Coordinate-wise Median (Coord)}
This method computes the median value for the horizontal ($x$) and vertical ($y$) axes independently. It minimizes the sum of absolute differences (L1 loss) along each dimension. While more robust to outliers than the mean, it ignores the spatial correlation between the $x$ and $y$ coordinates.
\begin{equation}
    \hat{p}_{\text{coord}} = \left( \operatorname{median}(\{x_i\}_{i=1}^K), \ \operatorname{median}(\{y_i\}_{i=1}^K) \right)
\end{equation}

\textbf{Geometric Median (Geo)}
The Geometric Median (also known as the L1-median or spatial median) is a robust estimator of central tendency in Euclidean space. Unlike the Coordinate-wise Median, it minimizes the sum of L2 distances to all sample points. Since no closed-form solution exists, we approximate it using Weiszfeld's algorithm.
\begin{equation}
    \hat{p}_{\text{geo}} = \underset{p \in \mathbb{R}^2}{\operatorname{argmin}} \sum_{i=1}^{K} \| p - p_i \|_2
\end{equation}

\textbf{Medoid}
The Medoid is similar to the Geometric Median in that it minimizes the sum of pairwise distances, but with the constraint that the selected point must be one of the original candidates in $\mathcal{P}$. This ensures the output is a valid coordinate explicitly generated by the model, rather than a synthesized point.
\begin{equation}
    \hat{p}_{\text{medoid}} = \underset{p \in \mathcal{P}}{\operatorname{argmin}} \sum_{i=1}^{K} \| p - p_i \|_2
\end{equation}
\section{Method}
In this section, we elaborate on the design principles governing the reward functions in the COPC framework. Our objective is to decouple the optimization of Candidate Generation (Proposer) and Visual Discrimination (Critic), ensuring they co-evolve without interference.

\textbf{Proposer Accuracy Reward ($R_{acc}$)}
Standard mean-based aggregation methods (e.g., minimizing average distance) often compel the model to converge toward the geometric center of the target distribution to minimize global loss, leading to "safe" but imprecise predictions in the invalid background. To address this, we employ a Softmax-weighted aggregation strategy that prioritizes the best-performing candidate. The quality of each individual candidate $p_i$ is first quantified using a Gaussian kernel $s_i = \exp(-\frac{||p_i - p_{gt}||^2}{2\sigma^2})$ to map Euclidean distances into a bounded similarity score. Crucially, the final reward aggregates these scores via Softmax, formulated as $R_{acc} = \sum \text{Softmax}(s_i) \cdot s_i + \frac{1}{K}\sum r_{hit, i}$. This mechanism assigns disproportionately high weight to the optimal candidate while suppressing the influence of sub-optimal outliers. Consequently, it effectively mitigates the penalty for exploration, allowing the Proposer to "gamble" on diverse locations without being punished for "misses," provided at least one candidate precisely captures the target intent.

\textbf{Proposer Coverage Reward ($R_{cov}$)}
To strictly prevent mode collapse—where the model generates $K$ identical or highly clustered points to minimize variance—we formulate a coverage reward based on the geometric spread of the candidates. Instead of simple pairwise distances, we utilize the determinant of the covariance matrix $\Sigma$ of the generated coordinates, which geometrically represents the "volume" or area spanned by the candidate distribution. We map this determinant to a normalized range using a hyperbolic tangent function: $R_{cov} = \tanh(\sqrt{|det(\Sigma)| + \epsilon})$. The determinant acts as a robust proxy for spatial diversity, ensuring the candidates are not colinear or collapsed into a single pixel, while the $\tanh$ normalization ensures the reward signal remains stable within $[0, 1]$. This formulation prevents unbounded reward growth and ensures the auxiliary diversity objective does not overwhelm the primary accuracy gradient, effectively encouraging the model to explore the spatial search space during the early stages of training.

\textbf{Critic Top-1 Selection Reward ($R_{top1}$)}
While the ranking quality is important, the ultimate utility of the GUI agent depends solely on the precision of the final executed action. Therefore, we design the Top-1 Selection Reward $R_{top1}$ as a direct proxy for the system's end-to-end success rate. It is defined simply as the Gaussian quality score of the candidate selected as the highest rank by the Critic, i.e., $R_{top1} = s_{best}$. By explicitly rewarding the quality of only the top-ranked candidate $p_{best}$, this sparse but high-signal objective ensures that the optimization process remains grounded in the task's primary goal: precise execution. This prevents the model from optimizing for general ranking metrics at the expense of the specific precision required for the top-1 output, aligning the Critic's objective directly with the inference-time success criterion.

\textbf{Critic Ranking Reward ($R_{ndcg}$)}
To cultivate fine-grained visual discrimination capability, we treat the Critic's task as a list-wise information retrieval problem rather than a simple binary classification task. Since candidate quality is continuous (based on distance to the ground truth) rather than binary, simple accuracy metrics are insufficient. We employ Normalized Discounted Cumulative Gain (NDCG), calculated as the ratio of Discounted Cumulative Gain (DCG) to Ideal DCG: $R_{ndcg} = \frac{DCG_K}{IDCG_K}$. This formulation uses a logarithmic decay factor in the denominator ($\log_2(i+1)$), which imposes significantly heavier penalties for ranking errors at the top of the list compared to the bottom. This forces the Critic to be extremely discriminating about the high-ranking candidates that directly impact potential selection, while being more tolerant of sorting errors among the low-quality "noise" candidates in the tail of the distribution, thereby focusing the model's capacity on differentiating the target from the most plausible distractors.
\section{Evaluation Benchmarks}
To comprehensively evaluate the generalization ability and grounding precision of the COPC framework across different platforms (Desktop, Mobile, Web) and complex application scenarios, we selected 6 representative mainstream benchmarks for testing. This section details the specific characteristics and data composition of these evaluation benchmarks (including MMBench-GUI, ScreenSpot-Pro, UI-Vision, etc.).

\textbf{MMBench-GUI }is a multi-platform GUI agent benchmark covering 6 systems (Windows, macOS, Linux, iOS, Android, Web) with 8123 tasks. It uses a 4-level hierarchical evaluation and the EQA metric (balancing success and efficiency), identifies visual localization as a core bottleneck, and provides open resources to advance GUI agent research.

\textbf{ScreenSpot-Pro} is a benchmark for evaluating MLLMs' GUI grounding in high-resolution professional environments, featuring 1,581 instruction-screenshot pairs from 23 apps (5 industries, 3 OSes), authentic high-res full-screen images, expert annotations, and tiny target elements to reflect real-world professional complexity.

\textbf{UI-Vision} is the first open-source benchmark for desktop graphical user interfaces (GUIs), filling the gap left by web/mobile-focused tools. It covers 83 open-source apps across 6 domains, with dense annotations and 8k+ query-label pairs.

\textbf{ScreenSpot-v2} is a revised GUI grounding benchmark derived from the original ScreenSpot dataset, with approximately 11.32\% of incorrect or ambiguous annotations corrected—including spelling errors, ambiguous instructions, redundant samples, and mislabeled bounding boxes—while preserving the total number of test samples. 

\textbf{UI-E2I-Synth} is a GPT-4o-powered pipeline for GUI instruction grounding, generating 9.9M instructions with 1.6M cross-platform (Web/Windows/Android) screenshots. It balances element types, includes explicit/implicit instructions, aligns with real-device element-to-screen ratios.

\textbf{OS-WorldG} is a fine-grained GUI grounding benchmark featuring 564 carefully annotated samples, covering five core competencies (text matching, element recognition, layout understanding, fine-grained manipulation, and rejection handling) with 32 annotated UI element types; each sample is paired with paraphrased, software-expertise-free instructions, requiring an average of 0.5 person-hours for annotation.


\section{Case Study}
In this section we provide detailed case studies to illustrate the "Propose-then-Critic" process. The Visual Feedback figures visualize the candidate coordinates generated by the Proposer during the first inference round. These candidates are rendered as numbered markers overlaid on the original GUI screenshot. In these figures, the red bounding box denotes the Ground Truth region.

\newtcolorbox{mybox}[1]{
    title={#1}, 
    colback=white, 
    colframe=gray!70, 
    fonttitle=\bfseries, 
    boxsep=5pt, 
    left=5pt,right=5pt,top=5pt,bottom=5pt 
}

\newtcolorbox{fullwidthbox}[1]{
    title={#1},
    colback=white, 
    colframe=gray!70, 
    fonttitle=\bfseries,
    boxsep=5pt,
    left=5pt,right=5pt,top=5pt,bottom=5pt,
    titlerule=0.5pt
}
\begin{figure*}[htbp] 
    \centering
    \includegraphics[width=0.9\textwidth]{figure/images/122_original.png} 
     \caption{Example 1}

\end{figure*}

\begin{figure*}[ht]
    \begin{fullwidthbox}{Prompt}
\textbf{Question:}  Enter Sculpt mode to use Inflate brush to enlarge some specific areas.
    \end{fullwidthbox}
\end{figure*}

\begin{figure*}[ht]
    \begin{fullwidthbox}{Proposer Output}
\textbf{ Proposer Output}: The user wants to enter Sculpt mode and use the Inflate brush to enlarge specific areas of a 3D object in Blender. The current screenshot already shows that the user is in Sculpt Mode, as indicated by the active "Sculpt Mode" tab in the top menu and the "Inflate" brush selected in the toolbar. The core intent is to ensure proper tool activation and focus on shaping the mesh. Based on the interface, the most critical elements for this task are: 1) Ensuring Sculpt Mode is active (top menu), 2) Selecting the Inflate brush (left toolbar), and 3) Hovering or clicking on the object to apply the brush. The top-left area contains the Sculpt Mode tab; the left scrollbar contains the tools including Inflate; and the 3D view center around the object is where the user will perform sculpting. To cover potential interactions, I spread coordinates across the top-left (menu activation), toolbar (tool selection), and 3D canvas (object interaction), ensuring diversity while including the Inflate brush and likely object interaction zone.

{
  "candidates": [
    [45, 32],
    [13, 233],
    [320, 452],
    [249, 32],
    [262, 430]
  ]
}

    \end{fullwidthbox}
\end{figure*}
\begin{figure*}[htbp] 
    \centering
    \includegraphics[width=0.9\textwidth]{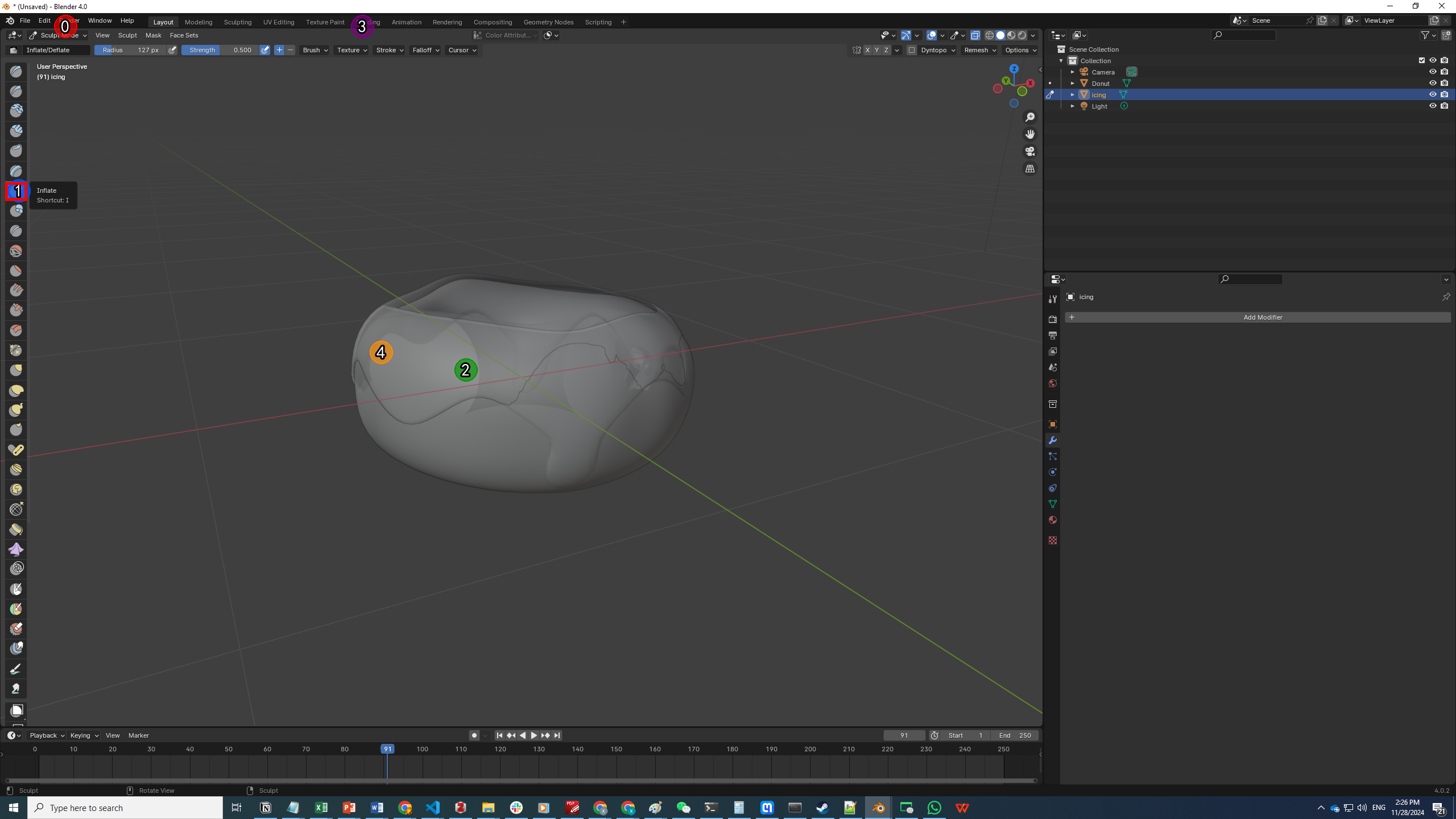} 
     \caption{Example 1 Visual Feedback}

\end{figure*}
\begin{figure*}[ht]
    \begin{fullwidthbox}{Critic Output}
\textbf{Critic Output}:[Analysis of candidates...]
The user's instruction is to enter Sculpt mode and use the Inflate brush to enlarge specific areas. The current screenshot already shows that the user is in Sculpt Mode. However, to ensure perfect alignment with the instruction, the top priority is making sure the "Inflate" brush is selected — which is visibly highlighted on the left toolbar. The candidate at ID 1 is pinpointing the "Inflate" brush icon itself, which is the primary target for fulfilling the user’s instruction. Candidates 0 and 3 are on the top menu bar — ID 0 is on the "Sculpt" menu, which is already active, and ID 3 is on the "Texture Paint" menu, which is irrelevant here — they are off-target but close to the sculpting context. Candidate 4 is on the 3D view outside the object, which is too far from the actual sculpting focus. Candidate 2 is on the object surface itself; this may be relevant for future sculpting, but per the instruction’s focus on tool activation, it is second-tier. Id 1 (Inflate brush) is most accurate and needs to be ranked \#1. Among candidates that are technically on-target (only ID 1 qualifies here), it is closest to visual center of its target. Thus, ID 1 is clearly the best. Subsequently, because ID 0 and ID 3 are on the top menu bar (relevant to mode), and ID 2 (surface interaction) likely aids the sculpting itself — the role of ID 2 purely as a “hitpoint” — the registrations go ranks 2, 3, 4. We need to include "3T:" and "S-4 blend" as contexts, but focus on core matching. Finally, since ID 4 is on the object surface, physically far from element useful to activate - particularly scores low. NX-6 (ETER-level ranked as 5points from Bear\_Jargon attribute selection) needs explicit ordering. Or: descending value.
json
\{
  "ranked\_ids": [1, 0, 3, 2, 4]
\}

    \end{fullwidthbox}
\end{figure*}

\begin{figure*}[htbp] 
    \centering
    \includegraphics[width=0.9\textwidth]{figure/images/290_original.png} 
     \caption{Example 2}

\end{figure*}

\begin{figure*}[ht]
    \begin{fullwidthbox}{Prompt}
\textbf{Question:}  Enter Sculpt mode to use Inflate brush to enlarge some specific areas.
    \end{fullwidthbox}
\end{figure*}

\begin{figure*}[ht]
    \begin{fullwidthbox}{Proposer Output}
\textbf{ Proposer Output}: The user wants to close a popup in Excel — specifically, the "Shape data with the Power Query Editor" notification that appears in the top-left corner. This is a common UI element that users want to dismiss quickly. The popup is anchored to the top-left, containing a green "Got it" button at its bottom-right corner, which is the primary affordance for closing it. Given standard UI design, the button’s location is predictable. The popup is small compared to the rest of the interface, so its approximate bounding box spans from roughly [5, 100] to [100, 180]. To cover plausible clicks: I’ll sample points around the button area (bottom-right of the popup) while also including points near the popup header (top-left) and middle-left area for ambiguity. This ensures diversity while targeting the closing mechanism.

\{
  "candidates": [
    [35, 120],
    [80, 160],
    [90, 185],
    [40, 140],
    [75, 110]
  ]
\}

    \end{fullwidthbox}
\end{figure*}
\begin{figure*}[htbp] 
    \centering
    \includegraphics[width=0.9\textwidth]{figure/images/290_add_point_with_gt.png} 
     \caption{Example 2 Visual Feedback}

\end{figure*}
\begin{figure*}[ht]
    \begin{fullwidthbox}{Critic Output}
\textbf{Critic Output}:[Analysis of candidates...]
The user's instruction is to "close the popup in Excel," and the popup in question is the "Shape data with the Power Query Editor" notification in the top-left of the Excel window. The only actionable element for closing this popup is the green "Got it" button at its bottom-right.

Candidate 0 (at [35, 120]) is in the top-left text area, not on the button.
Candidate 1 (at [80, 160]) is in the main body text, off-target.
Candidate 2 (at [90, 185]) is on the green "Got it" button — HIT.
Candidate 3 (at [40, 140]) is in the text area, off-target.
Candidate 4 (at [75, 110]) is in the header of the popup, off-target.

Since only candidate 2 is on the target (the "Got it" button), per Priority Rule 1, it is the absolute best. The other candidates are off-target. According to Rule 3 (for off-target), the one closest to the boundary of the button (which candidate 2 is-on) is already the best. The rest are farther away from the button region. No other candidate hits the button. Therefore, the ranking is clear: candidate 2 is best, and the others are progressively worse due to being farther away (and none are on-target).

```json
\{
  "ranked\_ids": [2, 0, 1, 3, 4]
\}
```

    \end{fullwidthbox}
\end{figure*}
\section{Information About Use Of Ai Assistants}
We used large language models (ChatGPT, Copilot) to assist with coding and linguistic polishing. No AI tools were used to generate the scientific ideas or novel contributions of this work.

%% file: tables/hyperparameters.tex
\begin{table}[!ht]
    
    \centering
    \setlength{\arrayrulewidth}{0.5pt}
    \resizebox{0.45\textwidth}{!}{
    \begin{tabular}{c|c}
    \toprule[2pt]
    \textbf{Hyperparameter} & \textbf{Setting} \\
    \midrule
    \midrule
    
     \cellcolor{gray!15}Clip Ratio & \cellcolor{gray!15}0.25 \\
     Top P & 0.99 \\
     \cellcolor{gray!15}Rollout Repeat Size & \cellcolor{gray!15}8 \\
     Temperature & 1.0 \\
     \cellcolor{gray!15}Optimizer & \cellcolor{gray!15}Adam \\
     KL Type & Low Var KL \\
     \cellcolor{gray!15}Learning rate & \cellcolor{gray!15}5e-7 \\
     Epochs & 2 \\
    \cellcolor{gray!15} momentum coefficient $\alpha$ & \cellcolor{gray!15}0.01 \\
    $\sigma$ & 0.5 * GT Box Width \\
        \cellcolor{gray!15} train batch size & \cellcolor{gray!15} 64 \\
    \bottomrule[2pt]
    \end{tabular}
    }
    \caption{Hyperparameters of Qwen3-VL-8B.}
    \label{tab:hyperparameters_and_training_schedule_large}
\end{table}